\documentclass{SciPost}

\hypersetup{
    colorlinks,
    linkcolor={red!50!black},
    citecolor={blue!50!black},
    urlcolor={blue!80!black}
}

\usepackage[bitstream-charter]{mathdesign}
\usepackage[utf8]{inputenc} % allow utf-8 input
\usepackage[T1]{fontenc}    % use 8-bit T1 fonts
\usepackage{amsmath}
\usepackage{xr-hyper}
\usepackage{hyperref}       % hyperlinks
\usepackage{subcaption}  
\usepackage{placeins}
\usepackage{longtable}
\usepackage{url}            % simple URL typesetting
\usepackage{booktabs}       % professional-quality tables
\usepackage{nicefrac}       % compact symbols for 1/2, etc.
\usepackage[numbers,sort&compress]{natbib}
\usepackage{microtype}      % microtypography
\usepackage{lipsum}
\usepackage{amsmath}
\usepackage{graphicx}
\usepackage{xcolor}
\usepackage{multirow}
\usepackage[capitalise]{cleveref}
\graphicspath{ {./images/} }
\DeclareSymbolFont{usualmathcal}{OMS}{cmsy}{m}{n}
\DeclareSymbolFontAlphabet{\mathcal}{usualmathcal}
\fancypagestyle{SPstyle}{
\fancyhf{}
\lhead{\colorbox{scipostblue}{\bf \color{white} ~SciPost Physics }}
\rhead{{\bf \color{scipostdeepblue} ~Submission }}

\fancyfoot[C]{\textbf{\thepage}}
}

\begin{document}

\pagestyle{SPstyle}

\begin{center}{\Large \textbf{\color{scipostdeepblue}{Local Diagnostics of Learned Event Representations for Anomaly Detection at the HL-LHC\\}}}\end{center}

\begin{center}\textbf{
Donatella Genovese\textsuperscript{1$\star$},
Philip Ploner\textsuperscript{2},
Simone Scardapane\textsuperscript{3},
Stefano Giagu\textsuperscript{4} and
Thea Klaeboe Aarrestad\textsuperscript{2}
}\end{center}

\begin{center}
{\bf 1}   Department of Computer, Control and Management Engineering, Sapienza University of Rome \\
{\bf 2}   Department of Physics, ETH Zurich \\
{\bf 3}   Department of Information Engineering, Electronics and Telecommunications, Sapienza University of Rome \\
{\bf 4}   Department of Physics, Sapienza University of Rome
\\[\baselineskip]
$\star$ \href{mailto:email1}{\small donatella.genovese@uniroma1.it}
\end{center}

\section*{\color{scipostdeepblue}{Abstract}}
\textbf{\boldmath{Anomaly detection, the identification of collision events that deviate from Standard Model expectations, is one of the most promising model-agnostic strategies for new-physics searches at the Large Hadron Collider. Learning powerful low-dimensional representations in which rare events are sought removes the need to commit to hand-crafted observables a priori, but at the cost of the immediate physical interpretation those observables provide. For anomaly detection to form part of a discovery pipeline, flagged events must therefore be diagnosed locally: both to distinguish potential new physics from mismodeling or detector effects, and to identify observables for targeted follow-up searches. We evaluate supervised and self-supervised neural embedding strategies on COLLIDE-2V, a 750-million-event dataset simulated under realistic High-Luminosity LHC conditions, using the reconstruction error of an autoencoder trained in the learned latent space as the anomaly score.  Across the configurations and benchmark signals considered, supervised objectives, which exploit simulation process labels during training, generally outperform their self-supervised counterparts and, in their best configurations, improve on an autoencoder trained directly on kinematic features, raising the signal efficiency at $10\%$ background acceptance on all held-out processes considered. We further introduce a local diagnostic pipeline that fits a Gaussian Mixture Model to the SM latent space, profiles each component through high-level observables without requiring process labels, localizes the flagged events, and ranks observables by their local discriminating power. The ranking recovers the defining kinematics of the held-out processes, and correlating the anomaly score with the same observables quantifies the extent to which the score tracks them, revealing cases in which the most discriminating observable is not the one the score follows most closely.
}}

\vspace{\baselineskip}

%%%%%%%%%% BLOCK: Copyright information
% This block will be filled during the proof stage, and finilized just before publication.
% It exists here only as a placeholder, and should not be modified by authors.
\noindent\textcolor{white!90!black}{%
\fbox{\parbox{0.975\linewidth}{%
\textcolor{white!40!black}{\begin{tabular}{lr}%
  \begin{minipage}{0.6\textwidth}%
    {\small Copyright attribution to authors. \newline
    This work is a submission to SciPost Physics. \newline
    License information to appear upon publication. \newline
    Publication information to appear upon publication.}
  \end{minipage} & \begin{minipage}{0.4\textwidth}
    {\small Received Date \newline Accepted Date \newline Published Date}%
  \end{minipage}
\end{tabular}}
}}
}
%%%%%%%%%% BLOCK: Copyright information

% For convenience during refereeing we turn on line numbers:
%\linenumbers

\vspace{10pt}
\noindent\rule{\textwidth}{1pt}
\tableofcontents
\noindent\rule{\textwidth}{1pt}
\vspace{10pt}

% Write your article contents here, starting from first \section.
% An example structure is given below.

\section{Introduction}
\label{sec:introduction}

The Large Hadron Collider (LHC) at CERN collides protons at the energy frontier to probe the fundamental structure of matter and to search for phenomena beyond the Standard Model (SM) of particle physics. Anomaly Detection (AD), which consists of flagging events that are improbable under a reference model of normality, typically the SM-only hypothesis or, in practice, its dominant background process, is one of the central strategies for uncovering such new physics. A central question in AD on collider data is the choice of representation: each event is described by a very high-dimensional set of measurements, and a compact, informative set of variables must be chosen or learned in which departures from the SM can be sought. In the traditional collider-data analysis pipeline, the raw detector readout, consisting of millions of sensor measurements per collision event, is processed through reconstruction algorithms that identify individual particles and estimate their physical properties. These reconstructed objects are subsequently condensed into a set of physically meaningful high-level quantities, such as particle energy, momentum, invariant masses, and direction. AD methods are then applied in this feature space, searching for events or populations of events that deviate from the expectations of SM background processes \cite{Belis2024}. Although this feature-based pipeline has been highly successful~\cite{ATLAS:2012yve, CMS:2012qbp}, its performance is inherently limited by the quality and completeness of the hand-crafted features used to represent each event. Recent approaches therefore seek to learn event representations directly from lower-level detector or particle information, replacing manually designed observables with data-driven embeddings optimized to capture the underlying structure of collider events \cite{Cerri2019}, following a broader trend in High Energy Physics (HEP), where representation learning is increasingly used to build low-dimensional latent spaces from particle- and detector-level information~\cite{Park, bardhan} that can be transferred across multiple downstream tasks \cite{Birk}.

In the context of AD, self-supervised and foundation-model representations have been shown to enhance the sensitivity of searches for physics beyond the SM ~\cite{Li, Dillon}, with contrastive and physics-informed objectives used to learn latent spaces tailored to AD ~\cite{Favaro, Dillon2}. By compressing collider events into structured embedding spaces, these representations also open the possibility of studying anomalies geometrically, relating anomaly scores to specific regions of the latent space. Connecting that geometry back to physical observables, however, remains largely unexplored, so the physical interpretation that hand-crafted observables provide by construction is not recovered.

In this work, we study how physically meaningful information can be recovered once an anomaly detector built on a learned representation has flagged a set of collision events, with the aim of understanding the origin of the anomalous response and of designing a targeted follow-up analysis. This requires locating the flagged events in the latent space and identifying the SM populations that surround them, determining which physical observables distinguish them from this local background, and establishing which observables the anomaly score is associated with.

We carry out this study on \textsc{Collide-2V} \cite{COLLIDE2V}, a 750-million-event dataset designed for the High-Luminosity LHC (HL-LHC) that combines large scale, a broad set of SM processes, and realistic detector conditions. Unlike smaller and more specialized benchmarks commonly used in AD studies, \textsc{Collide-2V} enables representation learning in the large-data regime for which modern contrastive and self-supervised methods are designed. This setting allows us to study not only AD performance, but also the structure and interpretability of the learned latent spaces at a scale relevant for future collider analyses.

We work at the event level using the full-reconstruction (offline) view (Full-Reco view), which provides analysis-quality particle-flow objects, jets, leptons, photons, and missing transverse energy. Event representations are learned with a Transformer encoder and physics-motivated augmentations~\cite{Dillon2}, using two objectives, SimCLR~\cite{simclr} and VICReg~\cite{vicreg}, each in a self-supervised and a supervised variant. The two variants differ mainly in the information used during training: the self-supervised one relies on augmentations alone and could in principle be trained directly on collision data, while the supervised one also uses the process labels of the simulated SM backgrounds, available by construction in simulation, to group events by process. In both cases no signal is seen during training, and the processes used to evaluate AD enter only at inference, so the search remains agnostic to the signal hypothesis. AD is then performed in the learned latent space with an autoencoder trained on the dominant SM background, whose reconstruction error serves as the anomaly score.

Our work makes three main contributions. First, under a shared Transformer backbone, we compare supervised and self-supervised event representations on \textsc{Collide-2V} and find that, among the objectives considered, SM process labels alone, without any information on the signal, yield the most informative latent spaces, as measured by linear-probe quality. Second, we show that AD in these supervised latent spaces, in their best-performing configurations, improves the signal efficiency over an autoencoder trained directly on kinematic features on all held-out processes considered. Third, and central to this work, we introduce a lightweight local diagnostic framework that processes the embeddings, a per-event anomaly score, and a set of observables. It partitions the SM latent space with a Gaussian Mixture Model (GMM), characterizes each component through high-level observables without process labels, assigns the flagged events to components, and ranks observables both by how well they separate these events from the local SM background and by how strongly they correlate with the anomaly score. Applied to $HH\to 4b$ and to a hidden-valley $Z'$, the framework recovers their defining physics, heavy-flavour content and lepton multiplicity respectively. For the $Z'$, however, lepton multiplicity is the strongest local discriminant but not the observable most correlated with the anomaly score.

The remainder of the paper is organized as follows. \cref{sec:related} introduces the related work in Representation Learning, AD and Interpretability in HEP. \cref{sec:methods} introduces the \textsc{Collide-2V} dataset and describes our methodology, covering the contrastive embedding framework, the anomaly-detection setup, and the local diagnostic framework. \cref{sec:results} presents and discusses the experimental results, and \cref{sec:conclusion} summarizes our findings.

\section{Related work}\label{sec:related}

\subsection{Representation Learning in High-Energy Physics}
The need to learn compact representations of collider data directly from low-level inputs, rather than relying on hand-crafted observables, has motivated a growing body of work on neural embedding methods in HEP. Several paradigms have been explored in this context, each differing in the training objective used to structure the latent space. Metric learning approaches have been applied at the jet level to learn geometrically meaningful embeddings of jet constituents \cite{Park}. Masked particle modeling, inspired by masked language models, trains encoders to predict dropped constituents from the remaining ones, and has been studied on jet-level datasets such as JetClass \cite{Golling}. Joint embedding predictive architectures have been explored for jet classification and tagging tasks, again working on jet constituents \cite{bardhan}. A systematic comparison of these paradigms (contrastive, masked, and generative) within a unified framework and shared transformer encoder was recently carried out on JetClass \cite{chen}, providing evidence that the choice of training objective significantly affects the structure and transferability of the resulting representation.

Among representation learning objectives, contrastive methods have received particular attention in HEP due to their ability to encode physical symmetries through augmentation design. Dillon et al. \cite{JetCLR} train a Transformer encoder with a self-supervised SimCLR loss \cite{simclr} and physics-motivated augmentations encoding known kinematic symmetries, and \cite{Dillon} apply this approach to QCD dijet events from the LHC Olympics Dataset \cite{LHCO}, producing a low-dimensional latent space for downstream AD. The most systematic comparison of contrastive strategies to date at the event level is \cite{Metzger}, evaluating supervised and self-supervised SimCLR as well as VICReg with both MLP and Transformer encoders on the ADC2021 dataset \cite{ADC2021}, and finding that larger embedding dimensionality favors classification via linear probing while smaller dimensionality improves AD performance.

A parallel line of research has aimed at building general-purpose foundation models for collider physics, pre-trained on large jet datasets and fine-tuned on downstream tasks. OmniJet-$\alpha$ \cite{Birk} demonstrated the first successful transfer between an unsupervised generation task and supervised jet tagging using a transformer architecture on jet constituents. HEP-JEPA extended this direction with a Joint-Embedding Predictive Architecture (JEPA) pre-trained on the 100M-jet JetClass dataset, achieving competitive performance on top tagging and quark-gluon discrimination ~\cite{bardhan}. While these works establish the viability of foundation models for jet-level tasks, they focus on classification and generation rather than AD, and operate at the substructure level rather than on full collision events. 

Using learned latent representations as the feature space for downstream AD has emerged as a promising strategy for improving sensitivity in model-agnostic searches. A contrastive latent space learned on LHC Olympics dijets has been shown to support weakly supervised bump hunting with performance comparable to searches run directly in physical space~\cite{Dillon}. Autoencoder-based anomaly scores have been evaluated on contrastive embeddings at the event level~\cite{Dillon2}, while supervised contrastive learning has been combined with the CATHODE density-estimation method on a diphoton dataset, showing that a signal-sensitive latent space can substantially improve weakly supervised discovery sensitivity~\cite{Li}. At the jet level, both the embedding norm and a normalized autoencoder reconstruction error in the latent space have been used as complementary anomaly scores~\cite{Favaro}.

\subsection{Anomaly Detection for New Physics Searches}
AD at the LHC can be broadly divided into two paradigms: overdensity estimation, which searches for localized excesses in kinematic distributions, and outlier detection, which identifies individual events unlikely under the SM hypothesis \cite{Belis2024}. The first paradigm encompasses weakly supervised methods that approximate the likelihood ratio between data and a background model. The CWoLa framework~\cite{CWoLa} exploits mixed signal and sideband regions to train a classifier without explicit signal labels, and has been deployed in experimental searches for generic dijet resonances~\cite{ATLAS_CWoLa}. ANODE~\cite{ANODE} and its successor CATHODE~\cite{CATHODE} replace the classifier with a conditional normalizing flow trained in data sidebands, which is then interpolated into the signal region to generate a background template; a classifier trained to distinguish data from this template approaches the optimal likelihood ratio. These methods have been systematically benchmarked in the LHC Olympics 2020 challenge~\cite{LHCO}, which provides a dijet dataset at the jet-constituent level, designed for resonant AD via bump hunt in the dijet invariant mass distribution.

The second paradigm relies on unsupervised models, most prominently autoencoders, trained exclusively on SM background events and evaluated via reconstruction error as an anomaly score. Early work demonstrated the use of VAEs for new physics mining, working on low-level trigger data consisting of jets, leptons, and missing transverse energy~\cite{Cerri2019}. Subsequent studies extended this by incorporating normalizing flow layers in the latent space of a VAE to improve anomaly score quality, evaluated on the Dark Machines challenge dataset~\cite{Jawahar2022}, which provides a multi-channel event-level dataset of reconstructed objects (jets, leptons, and missing transverse energy) across twelve Beyond Standard Model (BSM) signal hypotheses~\cite{DarkMachines}, serving as a standard benchmark for model-agnostic outlier detection. A key challenge shared by all outlier detection methods is the tendency to produce anomaly scores correlated with kinematic variables of interest, which can introduce sculpting effects in bump-hunt searches~\cite{Belis2024}.

More recent work has explored AD directly in learned latent spaces, combining representation learning with downstream outlier detection. A semi-supervised approach is introduced in~\cite{ANTELOPE}, where a Particle Flow Network is trained on background and a reference signal to produce a permutation-invariant event-level embedding, which is then fed to an autoencoder for unsupervised anomaly scoring; the method is evaluated on the LHC Olympics dataset~\cite{LHCO}. Most closely related to our work, \cite{Metzger} evaluates contrastive embeddings like SimCLR as latent spaces for AD on the ADC2021 dataset~\cite{ADC2021}, an event-level dataset of reconstructed objects (MET, electrons, muons, and jets) in tabular format simulated with Pythia and Delphes at 13 TeV, finding that the choice of embedding dimensionality and training objective significantly affects downstream sensitivity.

\subsection{Representation Analysis and Model Diagnostics in High-Energy Physics}
Model characterization and interpretability have become increasingly important concerns in HEP, where deep learning models are deployed as discovery tools and must offer physical insights beyond raw performance metrics \cite{Belis2024}. Post-hoc attribution methods like SHAP \cite{Shap} have been applied to classifiers to identify dominant physical features \cite{Pezoa2023}, though systematic evaluations demonstrate that post-hoc feature rankings can vary significantly across architectures \cite{Khot2023}. To address these limitations, architectural designs increasingly embed interpretability directly into the network. Symmetry-preserving models like PELICAN~\cite{PELICAN} enforce Lorentz invariance by construction, while Graph Transformers with Mixture-of-Experts use attention maps and expert specialization to relate their decisions to physical observables, providing physics-grounded interpretability for complex ATLAS signatures such as displaced dark photons and supersymmetric signals \cite{Genovese2025, Genovese2026}.

In the context of AD, probing the structure of learned representation spaces remains an open direction. \cite{Dillon2021} show that structured latent spaces, such as Gaussian mixture priors in VAEs, improve latent organization for jet-image AD, with decoder weights reflecting physically meaningful jet prototypes. \cite{Hallin2025} construct a unified latent space of BSM models using contrastive learning, showing that distances in latent space correlate with phenomenological similarities. However, existing approaches rarely examine how collision events selected by an anomaly detector are localized within the learned background representation space, or how the detector's score correlates with kinematic variables in those specific latent regions.

\section{Methods}\label{sec:methods}
\begin{figure*}[t]
  \centering
  \includegraphics[width=\textwidth]{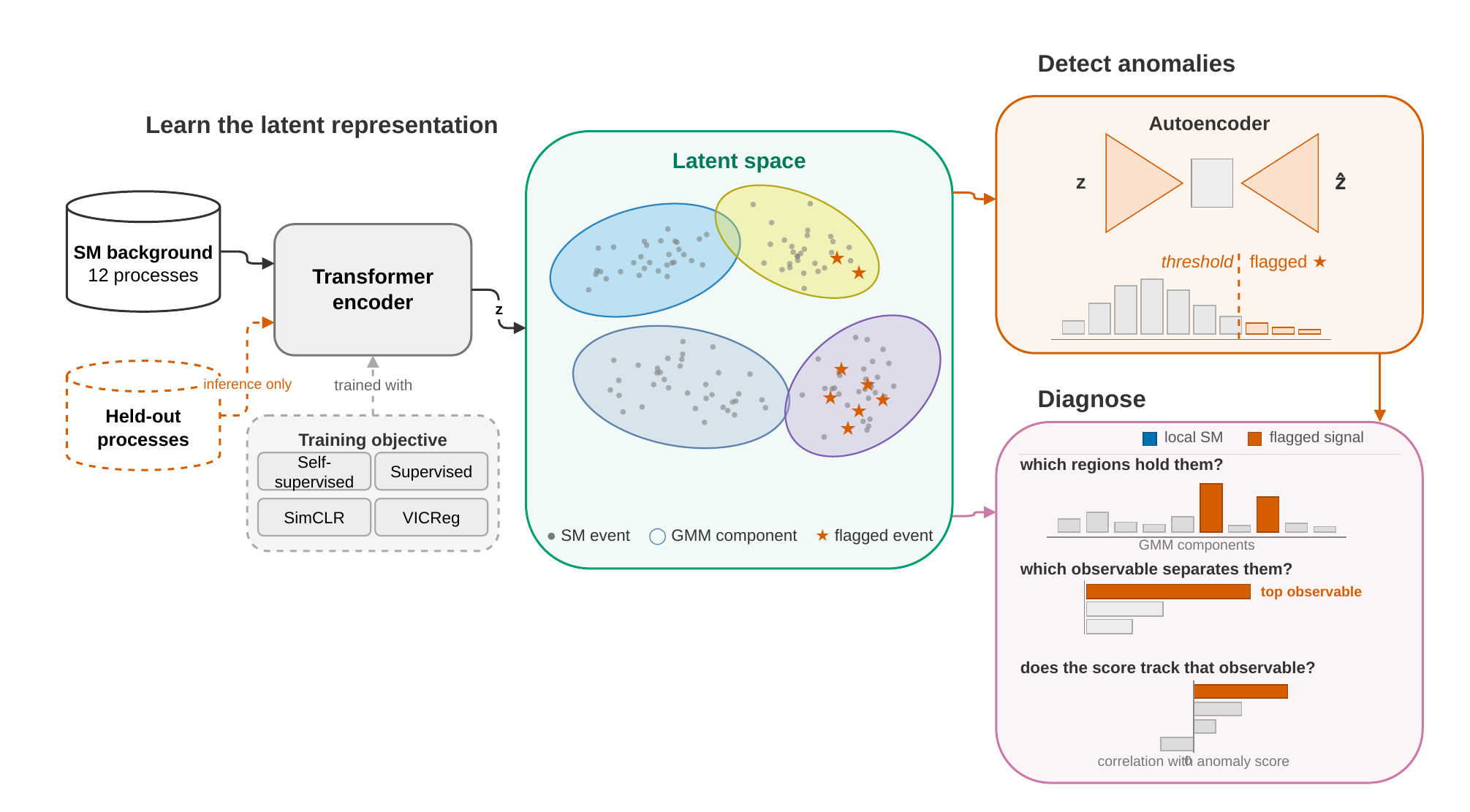}
  \caption{\textbf{End-to-end pipeline.} \emph{Left:} a Transformer encoder maps each event to a low-dimensional embedding $z$, trained on SM background processes; held-out events are passed at inference only. \emph{Center:} a GMM partitions the latent space into $K$ components. \emph{Top right:} an autoencoder flags anomalous events ($\star$) based on reconstruction error. \emph{Bottom right:} the diagnostic pipeline locates flagged events, ranks observables by local discriminating power, and tests their correlation with the anomaly score.}
  \label{fig:pipeline}
\end{figure*}

We present the methods that we use for our pipeline, summarized in \cref{fig:pipeline}. First, an encoder trained on SM events maps raw collisions from \textsc{Collide-2V} into a low-dimensional embedding space (\cref{sec:dataset}-\cref{sec:neuralemb}). An autoencoder trained in this space, on the dominant background alone, then assigns anomaly scores via reconstruction error (\cref{sec:adint}). Finally, a GMM is fitted in the same space, on the full SM population, providing a framework to characterize the space and analyze flagged events (\cref{sec:adint}).

\subsection{Dataset}
\label{sec:dataset}
We perform our study on \textsc{Collide-2V}~\cite{COLLIDE2V}, which provides 750 million HL-LHC collision events spanning a wide variety of processes under realistic detector conditions. Events are simulated using a standard three-stage pipeline: hard processes are generated with MadGraph~\cite{MG5}, followed by parton showering and hadronization with Pythia~\cite{Pythia}, and fast detector simulation with Delphes~\cite{Delphes} using a CMS Phase-2 HL-LHC card. Pileup is overlaid at the particle level with $\langle\mu\rangle = 200$, consistent with HL-LHC conditions.

We work at the event level using the Full-Reco view, which provides analysis-quality particle-flow objects reconstructed with the PUPPI pileup mitigation algorithm. Each event is represented as a set of reconstructed objects of five types, summarized in \cref{tab:objects}: up to 12 PUPPI AK4 jets, up to 8 electrons, up to 8 muons, up to 8 photons, and event-level missing transverse energy (MET). Variable-length object lists are zero-padded to their maximum multiplicity.

The processes used in this study are listed in \cref{tab:processes}: twelve SM processes covering the dominant production mechanisms at the HL-LHC, and two processes held out entirely from training and used to evaluate AD. $HH\to4b$ is a SM process treated as unknown, and serves as a signal proxy; $Z^{\prime} \to n(\mu\mu)$ is a BSM signal, in which a $1$~TeV hidden-valley $Z'$ decays to dark quarks that shower and hadronize in the hidden sector into dark mesons of mass $2$ GeV, each forced to decay to a muon pair. The dark meson multiplicity varies event by event, so $n$ denotes the resulting variable number of muon pairs. Further details on the dataset are provided in \cref{app:experimental}. Five further held-out processes, three additional proxies and two additional BSM signals, are used for the extended evaluation of \cref{app:signals}.

\begin{table}[ht]
\centering
\caption{Reconstructed objects in the Full-Reco view. Variable-length
         lists are zero-padded to the maximum multiplicity.}
\label{tab:objects}
\begin{tabular}{llcp{8cm}}
\toprule
\textbf{Object} & \textbf{Max.} & \textbf{\# features} & \textbf{Variables} \\
\midrule
AK4 jets (PUPPI) & 12 & 7 &
  $p_T$, $\eta$, $\phi$, mass, b-tag,
  $n_\text{charged}$, $n_\text{neutral}$ \\
Electrons & 8 & 8 &
  $p_T$, $\eta$, $\phi$, charge, isolation,
  $d_0$, $d_z$, $E_\text{had}/E_\text{em}$ \\
Muons & 8 & 7 &
  $p_T$, $\eta$, $\phi$, charge, isolation, $d_0$, $d_z$ \\
Photons & 8 & 5 &
  $p_T$, $\eta$, $\phi$, $E_\text{had}/E_\text{em}$, isolation \\
MET & -- & 2 &
  magnitude, $\phi$ \\
\bottomrule
\end{tabular}
\end{table}

\begin{table}[ht]
\centering
\caption{SM processes used to train the encoder, and the two held-out processes used to evaluate anomaly-detection performance.}
\label{tab:processes}
\begin{tabular}{llp{9cm}}
\toprule
\textbf{Type} & \textbf{\# processes} & \textbf{Process} \\
\midrule
\multirow{12}{*}{SM background}
  & \multirow{12}{*}{12}
  & QCD inclusive ($H_\mathrm{T}>50$\,GeV) \\
  & & QCD $\to b\bar{b}$ \\
  & & $W \to \ell\nu$ \\
  & & $W \to q\bar{q}$ \\
  & & $Z \to \nu\nu$ + jets \\
  & & $Z \to qq$ (uds) \\
  & & $Z \to bb$ \\
  & & $Z \to cc$ \\
  & & $\gamma$ \\
  & & $t\bar{t}$ (all-hadronic) \\
  & & $t\bar{t}$ (semi-leptonic) \\
  & & $t\bar{t}$ (all-leptonic) \\
\midrule
Signal proxy & 1 & $HH \to 4b$ \\
BSM signal   & 1 & $Z^{\prime} \to n(\mu\mu)$ \\

\bottomrule
\end{tabular}
\end{table}

\subsection{Neural embeddings}
\label{sec:neuralemb}
Representation learning describes the extraction of features using deep learning methods, where a neural network $\varphi_\theta$ parameterized by trainable parameters $\theta$ maps events from a high-dimensional input space $\mathcal{X}$ to a lower-dimensional latent space $\mathcal{Y}$: \begin{equation} \varphi_\theta : \mathcal{X} \longrightarrow \mathcal{Y}. \end{equation} 
The learned embeddings $z_i = \varphi_\theta(x_i)$ serve as the common input space for both AD and latent-space interpretability, described in \cref{sec:adint}. We consider four training configurations based on two loss functions (SimCLR~\cite{simclr} and VICReg~\cite{vicreg}) each in both a self-supervised and a supervised variant.

\paragraph{Self-supervised setting.}
In the self-supervised setting, no label information is available during training. The goal is to learn an encoder $\varphi_\theta$ that maps each event $x_i$ to a compact embedding $z_i = \varphi_\theta(x_i) \in \mathcal{Y}$ such that similar events are mapped to nearby regions of the latent space. Since labels are unavailable, the notion of similarity is defined through augmentations applied to the input events.

For SimCLR~\cite{simclr}, each event $x_k$ in a batch of $N$ events is augmented twice, producing a multiviewed batch of $2N$ samples $\{\tilde{x}_\ell\}_{\ell=1}^{2N}$, where $\tilde{x}_{2k-1}$ and $\tilde{x}_{2k}$ are two independently augmented views of $x_k$. The loss is computed for each sample in the multiviewed batch, which we refer to as the \textit{anchor}. For each anchor $i \in \mathcal{I} \equiv \{1, \ldots, 2N\}$, let $j(i)$ denote the index of the other augmented view originating from the same source event (the \textit{positive}) and let $\mathcal{A}(i) \equiv \mathcal{I} \setminus \{i\}$. The remaining $2(N-1)$ elements of $\mathcal{A}(i)$, i.e. the views originating from all other source events in the batch, are treated as \textit{negatives}. The SimCLR loss implements the attract-repel objective through a normalized temperature-scaled cross-entropy over the cosine similarity 
$\text{sim}(z_i, z_j) = z_i^\top z_j / \|z_i\|\|z_j\|$:
\begin{equation}
\mathcal{L}_{\text{SimCLR}} = \sum_{i \in \mathcal{I}} 
\ell_{\text{SimCLR}}(i), \quad
\ell_{\text{SimCLR}}(i) = -\log 
\frac{\exp\!\left(\text{sim}(z_i, z_{j(i)})/\tau\right)}
{\sum_{a \in \mathcal{A}(i)} 
\exp\!\left(\text{sim}(z_i, z_a)/\tau\right)},
\label{eq:simclr}
\end{equation}
where $\tau \in \mathbb{R}^+$ is a temperature hyperparameter. The denominator runs over all elements of $\mathcal{A}(i)$, so that the positive pair is normalized against the $2(N-1)$ negatives.

VICReg~\cite{vicreg} follows a different paradigm and does not rely on a multiviewed batch. Instead, it maintains two separate batches $Z = [z_1, \ldots, z_N]$ and $Z' = [z'_1, \ldots, z'_N]$, where $z_k$ and $z'_k$ are the embeddings of two independently augmented views of the same source event $x_k$. The loss combines three terms:
\begin{equation}
\mathcal{L}_{\text{VICReg}}(Z, Z') = \lambda\, s(Z, Z') 
+ \mu\left[v(Z) + v(Z')\right] 
+ \nu\left[c(Z) + c(Z')\right],
\label{eq:vicreg}
\end{equation}
where the invariance term $s(Z, Z') = \frac{1}{N}\sum_k 
\|z_k - z'_k\|^2_2$ attracts the two views of the same event using the Euclidean distance, and collapse prevention is achieved not through negative pairs but through two regularizers computed as global statistics over the batch. The variance term
\begin{equation}
v(Z) = \frac{1}{d_{\text{model}}}\sum_{j=1}^{d_{\text{model}}} \max\!\left(0,\, \gamma - 
S(z^j, \epsilon)\right),
\label{eq:var}
\end{equation}
where $z^j = [z_{1j}, \ldots, z_{Nj}]$ is the vector of values at dimension $j$ across the batch and $S$ is the regularized standard deviation, forces each embedding dimension to maintain high variance across the batch. The covariance term
\begin{equation}
c(Z) = \frac{1}{d_{\text{model}}}\sum_{i \neq j}[C(Z)]^2_{ij}, \quad 
C(Z) = \frac{1}{N-1}\sum_{k=1}^N(z_k - \bar{z})(z_k - 
\bar{z})^\top,
\label{eq:cov}
\end{equation}
penalizes the off-diagonal entries of the embedding covariance matrix, decorrelating the embedding dimensions and preventing informational collapse. 
\paragraph{Supervised setting.}
In the supervised setting, labels are available during training and are used to enrich the definition of positive pairs. For SimCLR, the key difference with respect to the self-supervised case is the definition of the positive set $\mathcal{P}(i)$ within the multiviewed batch: rather than containing only the augmented counterpart $j(i)$ of anchor $i$, it now includes all samples in the multiviewed batch sharing the same process label:
\begin{equation}
\mathcal{P}(i) = \{p \in \mathcal{A}(i) : \tilde{y}_p = 
\tilde{y}_i\},
\end{equation}
where $\tilde{y}_i$ denotes the process label of the $i$-th element in the multiviewed batch. All events sharing the same SM background label within the batch are therefore treated as positives, encouraging the latent space to organize events according to their underlying hard-scattering process. The supervised SimCLR (SupCon) loss then generalizes Eq.~\eqref{eq:simclr} by summing over all positives for each anchor, following~\cite{SupCon}:
\begin{equation}
\mathcal{L}_{\text{SupCon}} = \sum_{i \in \mathcal{I}} 
\ell_{\text{SupCon}}(i), \quad
\ell_{\text{SupCon}}(i) = 
\frac{-1}{|\mathcal{P}(i)|}
\sum_{p \in \mathcal{P}(i)}
\log 
\frac{\exp\!\left(\text{sim}(z_i, z_p)/\tau\right)}
{\sum_{a \in \mathcal{A}(i)} \exp\!\left(\text{sim}(z_i, 
z_a)/\tau\right)},
\label{eq:supcon}
\end{equation}
which recovers $\mathcal{L}_{\text{SimCLR}}$ when 
$|\mathcal{P}(i)| = 1$, i.e. when no other sample in the batch shares the same label as the anchor. In practice we train SupCon with an auxiliary classification head, adding a cross-entropy term to Eq.~\eqref{eq:supcon} with relative weight
$\lambda_{\mathrm{cls}}$:
\begin{equation}
\mathcal{L} = \mathcal{L}_{\mathrm{SupCon}} + \lambda_{\mathrm{cls}}\,\mathcal{L}_{\mathrm{CE}} .
\end{equation}

For supervised VICReg, we follow the Variance-Covariance Regularization (VCReg) approach of~\cite{VCReg}. In the supervised regime, a cross-entropy classification loss $\ell_{\text{sup}}$ is added to the objective, which already encourages similar representations for same-class events. The invariance term $s(Z, Z')$ of Eq.~\eqref{eq:vicreg} is therefore discarded, as its role is subsumed by the supervised loss. The variance and covariance regularizers are instead applied to the intermediate representations $h_i^{(j)} \in \mathbb{R}^{D_j}$, denoting the output of the $j$-th layer of $\varphi_\theta$ for input $x_i$, promoting feature diversity throughout the network. The training objective becomes:
\begin{equation}
\mathcal{L}_{\text{VCReg}}[\varphi_\theta] = 
\frac{1}{N}\sum_{i=1}^{N} \ell_{\text{sup}}(\varphi_\theta(x_i), 
y_i)
+ \sum_{j=1}^{M}\left[
\alpha\, \ell_{\text{var}}(h_1^{(j)}, \ldots, h_N^{(j)}) 
+ \beta\, \ell_{\text{cov}}(h_1^{(j)}, \ldots, 
h_N^{(j)})\right],
\label{eq:vcreg}
\end{equation}
where $\ell_{\text{sup}}$ is the cross-entropy loss, $M$ is the number of layers at which the regularization is applied, and $\alpha$, $\beta$ control the relative weight of the variance and covariance terms defined in Eqs.~\eqref{eq:var}--\eqref{eq:cov}. 

\paragraph{Physics-motivated augmentations.}
Event views are constructed from four augmentation operators, all acting on the preprocessed normalized feature vectors and applied online during training. \textit{Azimuthal rotation} applies a global rotation to all objects simultaneously, reflecting the exact rotational symmetry of the detector around the beam axis. \textit{Longitudinal boost} shifts all pseudorapidity values by a random displacement drawn from a Gaussian distribution, reflecting the approximate boost invariance of the hard process along the beam direction. \textit{Transverse momentum smearing} adds independent per-object Gaussian noise to the normalized transverse momentum of each object, modeling detector energy resolution effects. \textit{Random masking} zeroes parts of the event with a fixed probability $p_{\mathrm{mask}}$, simulating detector inefficiencies and missing particles, acting either on individual features or on entire objects.

These operators are combined into three mutually exclusive augmentation strategies. \textit{Physics} applies azimuthal rotation, longitudinal boost and $p_T$ smearing jointly to each view; \textit{Random feature} applies feature-level masking only; and \textit{Random particle} applies particle-level masking only. The strategy is treated as a per-model hyperparameter, and is used by SimCLR, SupCon and VICReg only: VCReg and the cross-entropy (CE) baseline operate on a single, un-augmented view, since in Eq. ~\eqref{eq:vcreg} the
invariance term is subsumed by the supervised loss and the remaining regularizers are computed on the intermediate representations of a single forward pass. The exact parametrization of each operator and the selection of the strategy are provided in \cref{app:experimental}.

\subsection{Anomaly Detection and Latent Space Diagnostics}\label{sec:adint}

\paragraph{Autoencoder-based anomaly detection.}
Following~\cite{Cerri2019, Dillon2}, AD is performed by training a simple fully-connected autoencoder on top of the learned embedding space. The autoencoder is trained exclusively on embeddings of the dominant QCD background, following common practice in outlier-based searches, where the most abundant and best-modeled process is used to define normality. The per-sample mean squared error (MSE) between input and reconstruction serves as the anomaly score:
\begin{equation}\label{eq:anomaly_score}
    s(z) = \frac{1}{d_{\mathrm{model}}}\,\|z - \hat{z}\|_2^2,
\end{equation}
where $\hat{z}$ is the reconstructed embedding and the average runs over the $d_{\mathrm{model}}$ embedding dimensions. An event is flagged as anomalous when $s(z) > s_{\mathrm{thr}}$, with $s_{\mathrm{thr}}$ calibrated on a held-out validation sample of QCD events to retain a background acceptance of 10\%. 

\paragraph{Latent-space interpretability via Gaussian Mixture Models.}
To characterize the structure of the SM embedding space and identify which regions correspond to anomalous events, we fit a GMM on the SM training embeddings. A GMM defines a distribution as a weighted sum of $K$ Gaussian components:
\begin{equation}
p(z) = \sum_{k=1}^{K} \pi_k \,\mathcal{N}(z \mid \mu_k, \Sigma_k),
\end{equation}
where $\pi_k \geq 0$ are the mixing weights with $\sum_k \pi_k = 1$, and $\mu_k$, $\Sigma_k$ are the mean and covariance of the $k$-th component. The parameters are estimated by  maximum likelihood via the Expectation Maximization algorithm~\cite{Dempster1977}. At each E-step, the responsibility of component $k$ for event $z_n$ is computed as the posterior probability:
\begin{equation}\label{eq:responsibility}
\text{r}(z_{nk}) = \frac{\pi_k \,\mathcal{N}(z_n \mid \mu_k, \Sigma_k)}
{\sum_{j=1}^{K} \pi_j \,\mathcal{N}(z_n \mid \mu_j, \Sigma_j)},
\end{equation}
and the M-step updates the component parameters using responsibility-weighted statistics. 

 The autoencoder defines anomalies relative to the QCD-inclusive reference distribution, whereas the GMM is fitted to the broader SM population to characterize the physical structure of the learned embedding space. Consequently, the signal efficiencies reported at fixed QCD acceptance quantify performance relative to this reference background, and should not be interpreted as efficiencies at the same acceptance for the complete SM background mixture. The diagnostic pipeline consists of five steps:

\begin{enumerate}
  \item \textbf{Partitioning.} A GMM with $K$ components is fitted on the SM
  training embeddings.
  \item \textbf{Label-free profiling.} Each component is characterized by the distributions of the observables of \cref{tab:gmm_variables} over the background events it contains, giving a physical baseline for each region without process labels.
  \item \textbf{Localization.} Signal events are assigned to the component of maximum posterior responsibility.
  \item \textbf{Local separation.} In the components holding the flagged
  signal, observables are ranked by the Wasserstein-1 distance~\cite{Givens:1984} between signal and all local SM events assigned to the same component, standardized by the standard deviation of the observable over the SM population. We report this against both the local SM population and the local QCD population.

  \item \textbf{Score-observable correlation.} Within the same components we compute the Spearman correlation between anomaly score and each observable, testing whether the score tracks the observable that separates the signal locally.
\end{enumerate}

The GMM construction and the physical profiling of its components (steps~1--2) do not require process labels. In the present benchmark study, the known identities of the simulated signal samples are used retrospectively to validate the diagnostic framework, by comparing the flagged signal events with the local SM background (steps~3--5). In an application to collision data, the corresponding comparison would involve the observed population of flagged events and an appropriate local background reference, without assuming prior knowledge of the physical origin of the anomalies.

\begin{table}[htbp]
  \centering
  \caption{High-level physical observables used to profile the GMM components and to rank the variables that separate flagged signal events from the local SM background. Exact definitions are given in \cref{app:observables}.}
  \label{tab:gmm_variables}
  \begin{tabular}{ll}
    \toprule
    \textbf{Observable} & \textbf{Physical meaning} \\
    \midrule
    $H_T$              & Hadronic activity \\
    MET                & Missing transverse energy \\
    $n_{\rm jets}$     & Jet multiplicity \\
    $n_{b\text{-tag}}$ & Heavy-flavour content \\
    $n_{\rm leptons}$  & Lepton multiplicity \\
    $M_{jj}$           & Leading dijet invariant mass \\
    $|\Delta\eta_{jj}|$  & Leading dijet pseudorapidity gap \\
    $M_T$              & Transverse mass \\
    \bottomrule
  \end{tabular}
\end{table}

\subsection{Experimental setup}
\label{sec:expsetup}
Having defined the embedding strategies (\cref{sec:neuralemb}), the AD and interpretability pipeline (\cref{sec:adint}), we now describe how these components are instantiated and evaluated jointly on \textsc{Collide-2V}. All models share the same Transformer encoder backbone, tokenizing each event by object type and processing it with a stack of Transformer encoder layers at a common model dimension $d_\text{model}$; for SupCon and SimCLR, the contrastive projection head is discarded after training, so linear probing and AD always operate on the $d_\text{model}$-dimensional encoder output. Full backbone, training, and dataset-split details are given in \cref{app:experimental}.

The encoder is trained on the 12 SM background processes of \cref{tab:processes} and optimized with AdamW~\cite{adamw}; representation quality is assessed via linear probing on a held-out SM test set. 
For AD, a fully-connected autoencoder is trained exclusively on QCD-inclusive embeddings and evaluated on the held-out processes of \cref{tab:processes}, reporting the area under the ROC curve (AUROC) and signal efficiency at a fixed 10\% background acceptance, both computed against QCD-inclusive as the background hypothesis. We scan $d_\text{model} \in \{32, 64, 128, 256\}$ for every model, with all other hyperparameters fixed as reported in \cref{app:ablation}; all results are mean $\pm$ std over five independent encoder seeds, one autoencoder per seed.

For the GMM interpretability pipeline, the mixture is fitted on the SM training embeddings of the selected representation; the embedding dimension, the number of components $K$, and the covariance structure are chosen as described in \cref{app:gmm_ablation}.

\section{Results}\label{sec:results}
\subsection{Comparison of embeddings}
\label{sec:results_probe}

\cref{tab:probe_auroc} reports the macro-averaged AUROC of a linear classifier trained on the frozen encoder embeddings across the 12 SM processes, for each model in its best configuration as selected by the ablation study of \cref{app:ablation}, conducted on a reduced training subset of the dataset. Results broken down by individual SM process are provided in \cref{app:addresults}. We also include a fully supervised model trained with a standard cross-entropy (CE) loss on the same Transformer backbone, with its own configuration reported in \cref{app:ablation}, and a linear classifier trained directly on the raw preprocessed kinematic features. Within the set of models considered, each supervised objective outperforms its self-supervised counterpart in linear-probe quality, with VCReg, SupCon, and CE clustering well above SimCLR and VICReg. 

\begin{table}[ht]
\centering
\caption{Macro-averaged AUROC (mean~$\pm$~std over 5 seeds) of a linear probe on the 12 SM classes, for each model in its best configuration (\cref{app:ablation}). For CE the AUROC is obtained from the classifier's own test predictions.}
\label{tab:probe_auroc}
\begin{tabular}{lcc}
\toprule
\textbf{Model} & $d_\text{model}$ & \textbf{AUROC} \\
\midrule
SupCon  & 128 & $0.9051 \pm 0.0020$ \\
SimCLR  & 256 & $0.8684 \pm 0.0014$ \\
VCReg   & 128 & $\mathbf{0.9075 \pm 0.0002}$ \\
VICReg   & 256 & $0.7682 \pm 0.0314$  \\
\midrule
CE (standard) & 256 & $0.9022 \pm 0.0004$ \\
Raw features  & --- & $0.8653 \pm 0.0007$ \\
\bottomrule
\end{tabular}
\end{table}

\subsection{Anomaly detection performance}
\label{sec:ad_results}
We assess anomaly-detection performance using the autoencoder-based pipeline introduced in \cref{sec:adint}: for each embedding model and dimension, a fully-connected autoencoder is trained exclusively on the embeddings of QCD-inclusive events, and its per event reconstruction error (\cref{eq:anomaly_score}) is used as the anomaly score. As a model independent baseline, we additionally train an autoencoder directly on the raw preprocessed kinematic features (Raw AE). Performance is evaluated on the two held-out processes of \cref{tab:processes} using two complementary metrics: the AUROC and the signal efficiency at a fixed background acceptance of 10\%, both reported as mean~$\pm$~std over 5 random seeds.

For readability we report here only the two supervised objectives, SupCon and VCReg. The self-supervised SimCLR and VICReg are reported in \cref{app:signals}, together with results on five further held-out processes.

\cref{tab:ad_signals} reports both metrics for each model and embedding dimension.
On $HH \to 4b$, both supervised objectives improve with $d_{\text{model}}$ and exceed the Raw AE baseline from $d_{\text{model}} = 64$, with VCReg at $d_{\text{model}} = 256$ reaching a signal efficiency of $97.2\%$ against $77.3\%$.
On $Z' \to n(\mu\mu)$, where the baseline already reaches $96.8\%$, every configuration except VCReg at $d_{\text{model}} = 32$ exceeds it.

\begin{table}[htbp]
\centering
\caption{Anomaly-detection performance on the two held-out processes for each model and embedding dimension: AUROC and signal efficiency at 10\% background acceptance (TPR @ FPR = 10\%), mean $\pm$ std over 5 seeds. Raw AE is trained directly on preprocessed kinematic features.}
\label{tab:ad_signals}
\begin{tabular}{ll cc cc}
\toprule
 & & \multicolumn{2}{c}{$HH\!\to\!4b$} & \multicolumn{2}{c}{$Z^{\prime} \to n(\mu\mu)$} \\
\cmidrule(lr){3-4}\cmidrule(lr){5-6}
\textbf{Model} & $d_\text{model}$ & \textbf{AUROC} & \textbf{TPR @ 10\% [\%]}
                                  & \textbf{AUROC} & \textbf{TPR @ 10\% [\%]} \\
\midrule
Raw AE & --- & $0.931 \pm 0.001$ & $77.3 \pm 0.7$ & $0.984 \pm 0.001 $ & $96.8 \pm 0.1$ \\
\midrule
\multirow{4}{*}{SupCon}
 & 32  & $0.892 \pm 0.031$ & $65.2 \pm 9.6$ & $\mathbf{0.996 \pm 0.001}$ & $\mathbf{99.0 \pm 0.2}$ \\
 & 64  & $0.937 \pm 0.017$ & $79.9 \pm 6.8$ & $0.995 \pm 0.001$ & $99.0 \pm 0.1$ \\
 & 128 & $0.943 \pm 0.005$ & $82.4 \pm 1.9$ & $0.995 \pm 0.001$ & $98.9 \pm 0.1$ \\
 & 256 & $0.954 \pm 0.009$ & $86.4 \pm 4.2$ & $0.995 \pm 0.001$ & $98.9 \pm 0.1$ \\
\midrule
\multirow{4}{*}{VCReg}
 & 32  & $0.937 \pm 0.025$ & $78.9 \pm 9.6$ & $0.977 \pm 0.015$ & $95.4 \pm 4.4$ \\
 & 64  & $0.937 \pm 0.012$ & $80.2 \pm 4.4$ & $0.991 \pm 0.002$ & $98.4 \pm 0.4$ \\
 & 128 & $0.959 \pm 0.005$ & $90.0 \pm 2.5$ & $0.991 \pm 0.004$ & $98.4 \pm 0.5$ \\
 & 256 & $\mathbf{0.982 \pm 0.004}$ & $\mathbf{97.2 \pm 1.7}$ & $0.991 \pm 0.001$ & $98.4 \pm 0.2$ \\
\bottomrule
\end{tabular}
\end{table}

\subsection{Localizing and explaining flagged anomalies}
\label{sec:results_gmm}

We apply the pipeline described in \cref{sec:adint} to the $HH \rightarrow 4b$ signal proxy and the $Z^{\prime} \to n(\mu\mu)$ signal, using the VCReg embedding with $d_{\text{model}} = 256$, a configuration that provides stable AD performance across both signals. As detailed in \cref{app:gmm_ablation}, the GMM is fitted on the SM training embeddings after reducing the dimensionality to 64 via Principal Component Analysis (PCA). The space is partitioned into $K = 7$ components, a value selected based on an occupancy criterion to prevent over-fragmentation and ensure each component captures meaningful data structures (\cref{app:gmm_ablation}).

\begin{figure}[tbp]
  \centering
  \includegraphics[width=\textwidth]{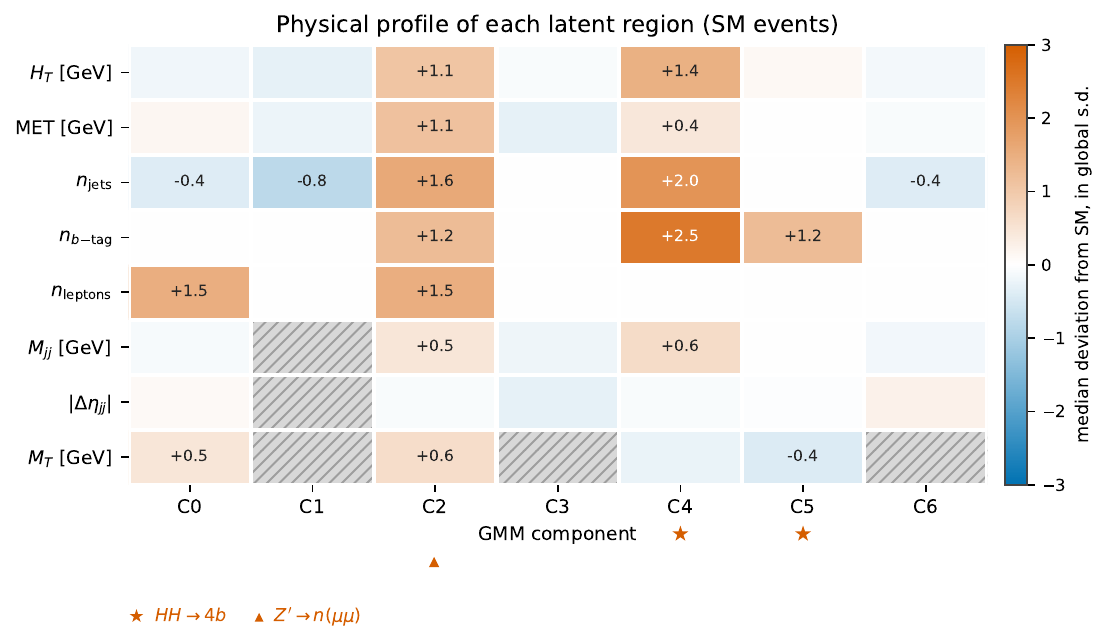}
\caption{Physical profile of each latent region, computed on the SM events assigned to it. The labels C0--C6 index the mixture components. Each cell displays the standardized median deviation: the median of the observable within the component minus the global SM median, divided by the observable's standard deviation over the SM population. Deviations are computed relative to the full SM population. For $n_{b\text{-tag}}$ and $n_{\text{leptons}}$ the global SM median is zero, so those rows cannot deviate downwards. Medians are used to ensure robustness against long distribution tails. Cells with absolute deviations smaller than 0.35 are left unlabeled to avoid visual clutter. Hatched cells are those where fewer than $15\%$ of the component's events yield a value.}
\label{fig:gmm_profiles}
\end{figure}

\cref{fig:gmm_profiles} characterizes each of the $K=7$ components through the distributions of the high-level observables of \cref{tab:gmm_variables}, computed on the SM events they contain. We describe each region using the \textit{standardized median deviation}; specifically, each cell displays the difference between the median of the observable within the component and its global SM median, divided by the observable's standard deviation over the SM population. Medians are used instead of means to remain robust against the long tails typical of kinematic distributions like $H_T$ and $M_{jj}$.

This description of the latent space is obtained entirely without process labels. For example, C4 emerges as a high-multiplicity, heavy-flavour region: the value of $+2.5$ for $n_{b\text{-tag}}$ indicates that events in C4 typically have 2 more b-jets than the global SM median, a shift equivalent to 2.5 standard deviations. Similarly, $n_{\text{jets}}$ is shifted by $+2.0$ standard deviations, showing that the normalization successfully places multiplicities and energies (like $H_T$, shifted by $+1.4$) on equal footing. C2 represents a distinct leptonic region with substantial hadronic activity and missing energy ($\mathbf{+1.5}$ for $n_{\text{leptons}}$, $+1.6$ for $n_{\text{jets}}$, and $\mathbf{+1.1}$ for MET). Conversely, C0 is leptonic but jet-poor, while C1 and C6 sit below the SM median in jet multiplicity and C3 shows no marked shift in any observable. Cells with absolute deviations smaller than 0.35 represent negligible shifts and are left unlabeled for visual clarity.

\begin{figure}[tbp]
  \centering
  \includegraphics[width=0.49\linewidth]{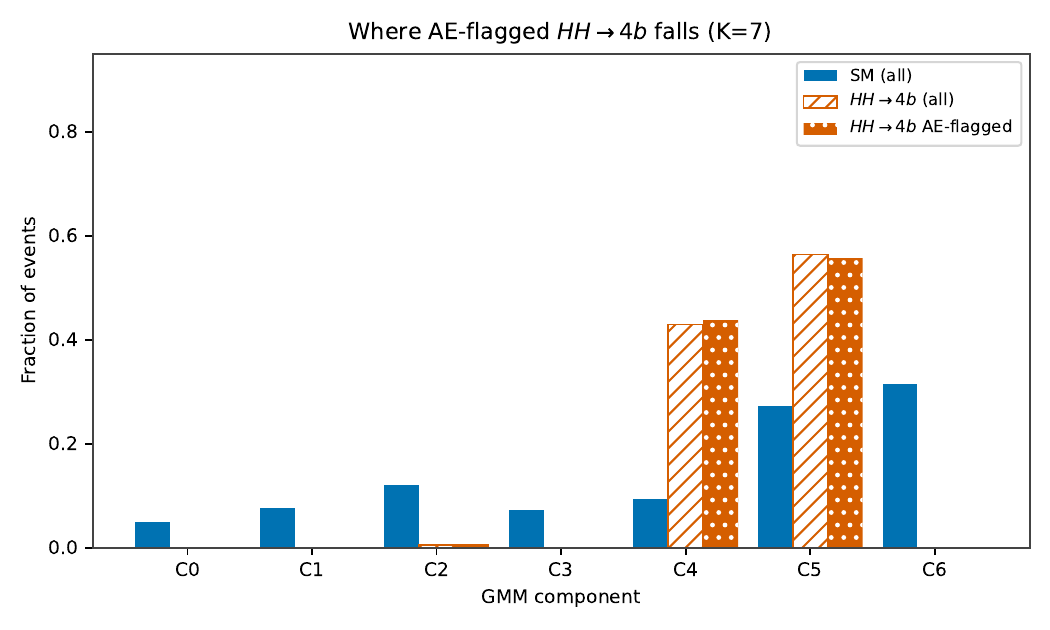}
  \hfill
  \includegraphics[width=0.49\linewidth]{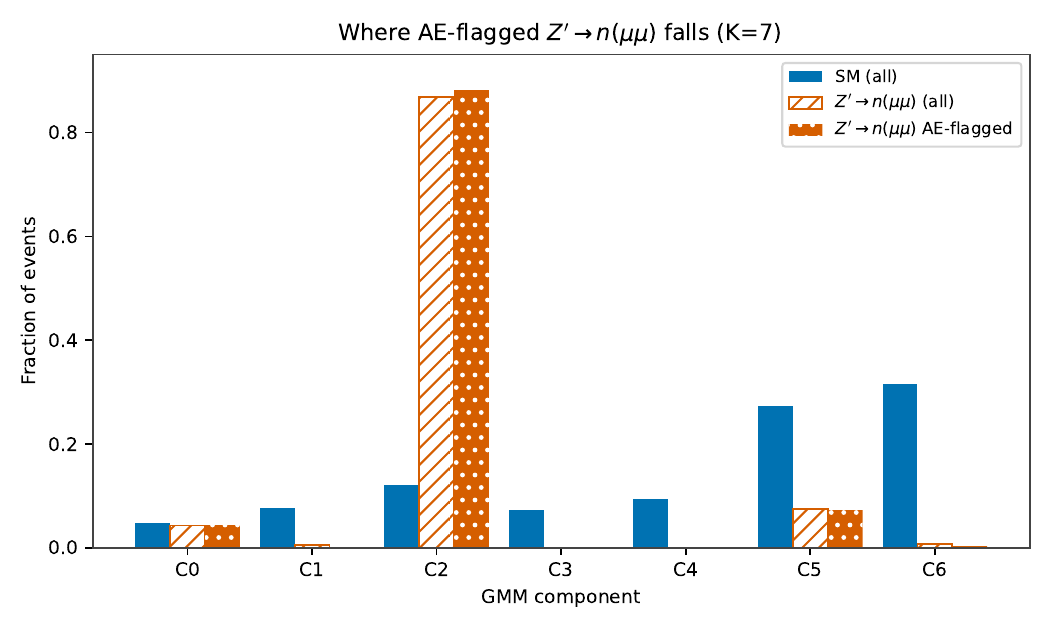}
  \caption{Assignment of the AE-flagged signal to the mixture components, for
  $HH\to4b$ (left) and $Z^{\prime} \to n(\mu\mu)$ (right), with the SM population
  shown for reference. Fractions are normalized independently for each population; the plotted occupancies therefore indicate localization in latent space rather than signal purity.}
  \label{fig:localisation}
\end{figure}

\cref{fig:localisation} shows the assignment of AE-flagged signal events to the GMM components, together with the SM background for reference. The flagged signals are highly localized, and their dominant components are disjoint. The $HH \rightarrow 4b$ primarily occupies C5 and C4, holding 55.6\% and 43.8\% of its flagged events, respectively, while 88\% of the flagged $Z^{\prime} \to n(\mu\mu)$ events fall into C2 alone.

Although localization reveals the broad kinematic profile of the anomalies, we investigate whether there are defining differences in the observable distributions between the flagged signal and the SM background mapped to that same region. \cref{tab:wasserstein_rank} ranks the observables using the Wasserstein-1 distance ($W_1^{\text{SM}}$) between the flagged signal and the local SM background within the dominant components. The local ranking correctly recovers the defining variables of the underlying physical processes: $n_{b\text{-tag}}$ is ranked first in C5 for $HH \to 4b$ ($W_1^{\text{SM}} = 2.06$), and $n_{\text{leptons}}$ is dominant in C2 for $Z' \to n(\mu\mu)$ ($W_1^{\text{SM}} = 8.61$).

\begin{table}[t]
\centering
\caption{Local discriminating power of the high-level observables within the mixture components that hold the AE-flagged signal, for $HH \rightarrow 4b$ (left) and $Z^{\prime} \to n(\mu\mu)$ (right). Only components holding at least $10\%$ of the flagged signal are reported. $W_1^{\text{SM}}$ is the Wasserstein-1 distance between the flagged signal and all SM events assigned to the same component, irrespective of their anomaly score, standardized by the standard deviation of each observable over the SM population. $W_1^{\text{QCD}}$, computed against the local QCD background (the autoencoder's trained normality), is reported as a control baseline. Observables are ranked by $W_1^{\text{SM}}$, with the top-ranked observable set in bold. C2 is essentially free of QCD, so the control cannot be computed there.}
\label{tab:wasserstein_rank}
\begin{minipage}[t]{0.48\textwidth}
\centering
\textbf{$HH \to 4b$}\\[2pt]
\begin{tabular}{llrr}
\toprule
& Observable & $W_1^{\mathrm{SM}}$ & $W_1^{\mathrm{QCD}}$ \\
\midrule
C5 & \textbf{$n_{b\text{-tag}}$} & \textbf{2.06} & 2.36 \\
 & $n_\mathrm{jets}$ & 0.65 & 0.34 \\
 & $H_T$ & 0.52 & 0.26 \\
 & $n_\mathrm{leptons}$ & 0.49 & 0.54 \\
  & $M_{jj}$ & 0.23 & 0.14 \\
 & $|\Delta\eta_{jj}|$ & 0.23 & 0.38 \\
 & MET & 0.18 & 0.05 \\
 & $M_T$ & 0.01 & 0.09 \\
\midrule
C4 & \textbf{$n_{b\text{-tag}}$} & \textbf{1.09} & 1.64 \\
 & $n_\mathrm{leptons}$ & 0.36 & 0.28 \\
 & MET & 0.22 & 0.25 \\
 & $n_\mathrm{jets}$ & 0.17 & 0.11 \\
 & $M_T$ & 0.07 & 0.13 \\
 & $H_T$ & 0.05 & 0.10 \\
 & $|\Delta\eta_{jj}|$ & 0.03 & 0.10 \\
 & $M_{jj}$ & 0.02 & 0.21 \\
\bottomrule
\end{tabular}
\end{minipage}
\hfill
\begin{minipage}[t]{0.48\textwidth}
\centering
\textbf{$Z^{\prime} \to n(\mu\mu)$}\\[2pt]
\begin{tabular}{llrr}
\toprule
& Observable & $W_1^{\mathrm{SM}}$ & $W_1^{\mathrm{QCD}}$ \\
\midrule
C2 & \textbf{$n_\mathrm{leptons}$} & \textbf{8.61} & --- \\
 & $M_T$ & 3.32 & --- \\
 & $n_{b\text{-tag}}$ & 1.68 & --- \\
 & MET & 1.57 & --- \\
 & $n_\mathrm{jets}$ & 1.37 & --- \\
 & $M_{jj}$ & 0.50 & --- \\
 & $H_T$ & 0.19 & --- \\
 & $|\Delta\eta_{jj}|$ & 0.05 & --- \\
\bottomrule
\end{tabular}
\end{minipage}
\end{table}

\begin{figure}[tbp]
\centering
\includegraphics[width=\textwidth]{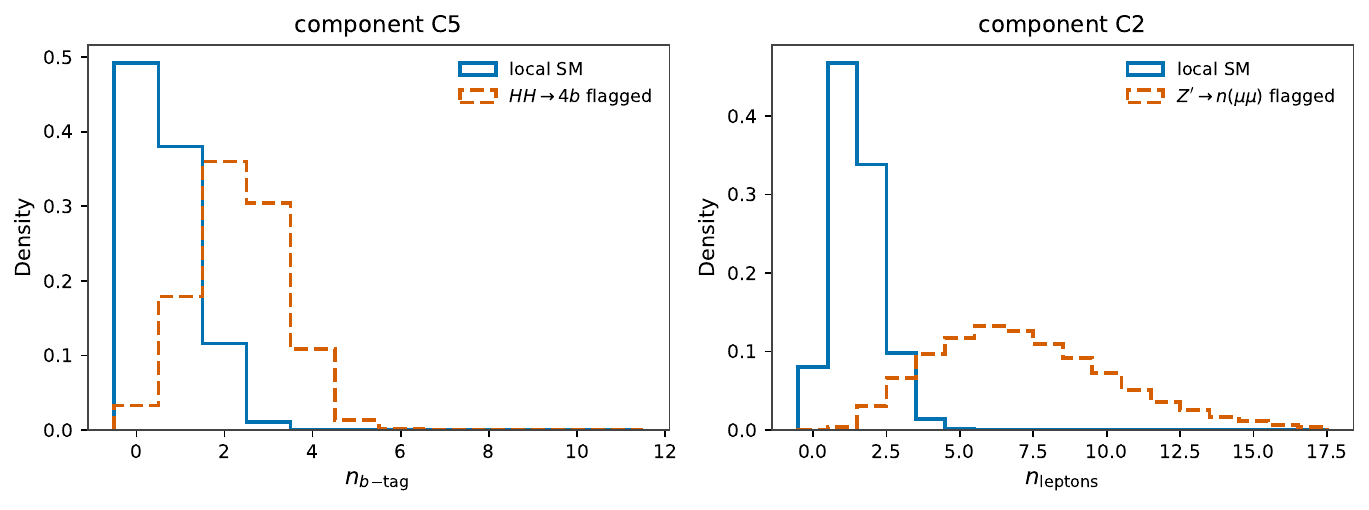}
\caption{Density distributions of the top-ranked observable in the dominant component for each flagged signal: $n_{b\text{-tag}}$ in C5 for $HH \rightarrow 4b$ (left) and $n_{\text{leptons}}$ in C2 for $Z^{\prime} \to n(\mu\mu)$ (right). The distributions compare the AE-flagged signal against the local SM background in the same latent region.}
\label{fig:top_observable}
\end{figure}

\cref{fig:top_observable} illustrates this separation by comparing the density distributions of these top-ranked variables for the flagged signals versus the local SM backgrounds. For example, in C5, the local SM background has a median of one b-jet, compared to two for the $HH \rightarrow 4b$ signal.

\begin{figure}[tbh!]
\centering
\begin{subfigure}[b]{0.8\textwidth}
    \centering
    \caption{Spearman correlation for $HH \to 4b$}
    \includegraphics[width=\textwidth]{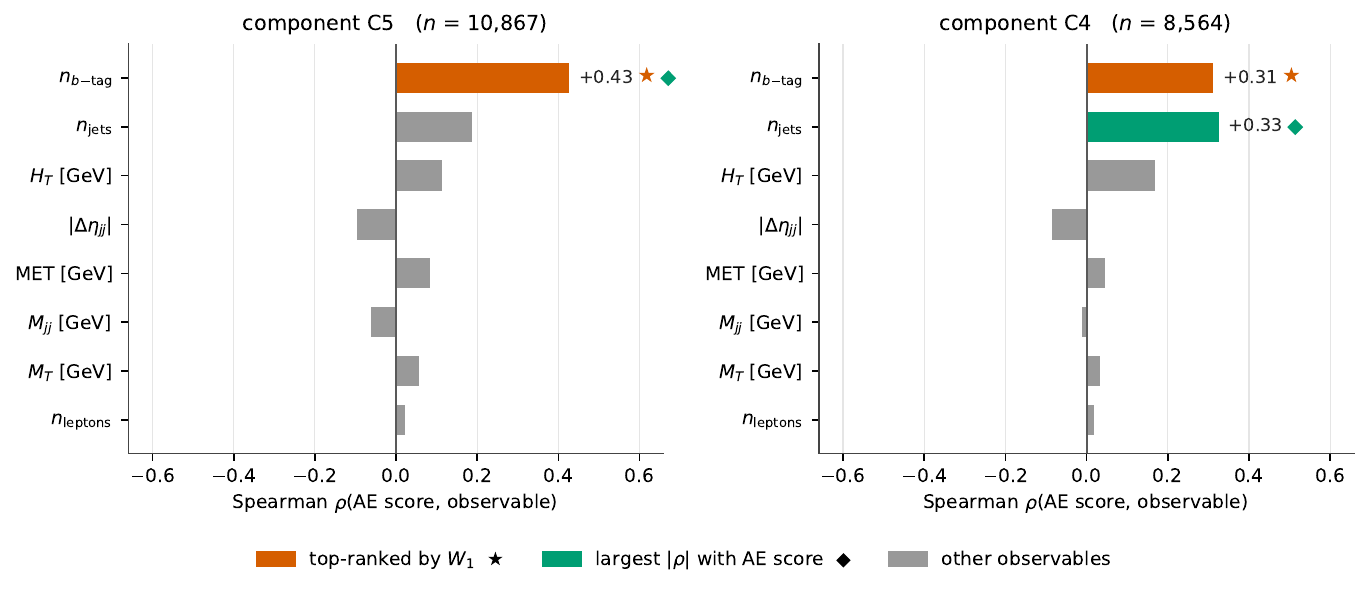}
    \label{fig:spearman_hh4b}
\end{subfigure}

\vspace{0.4em}

\begin{subfigure}[b]{0.4\textwidth}
    \centering
    \caption{Spearman correlation for $Z^{\prime} \to n(\mu\mu)$}
    \includegraphics[width=\textwidth]{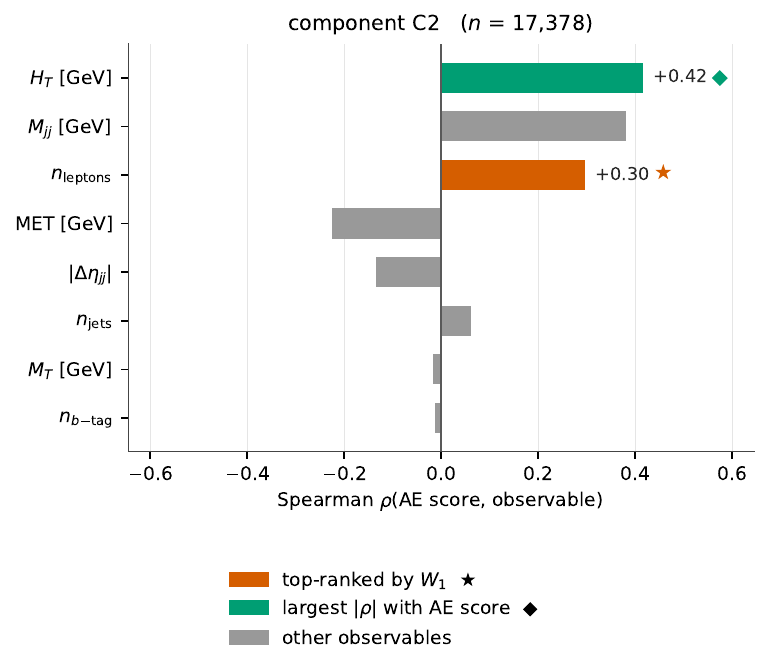}
    \label{fig:spearman_hvdilep}
\end{subfigure}

\caption{Spearman correlation between the autoencoder reconstruction error (anomaly score) and the physical observables for the flagged signal events, computed within their dominant GMM components. In each panel the observable ranked first by Wasserstein distance is marked in vermilion ($\bigstar$), and the one most strongly correlated with the score in green ($\blacklozenge$).}
\label{fig:spearman}
\end{figure}

Finally, to understand whether the autoencoder's decision aligns with the physical separation identified in \cref{fig:top_observable}, we compute the Spearman correlation between the AE reconstruction error (the anomaly score) and the observables for the flagged signal events. \cref{fig:spearman} displays these correlations within the dominant components for each signal. For the $HH \rightarrow 4b$ in C5, the anomaly score correlates most strongly with $n_{b\text{-tag}}$ ($\rho=0.43$), tracking the exact same observable selected by the local $W_1^{\text{SM}}$ ranking. In C4 the correlation is weaker ($\rho = 0.31$) and comparable to that of $n_{\text{jets}}$. This is consistent with the autoencoder responding to the heavy-flavour content of these events. Interestingly, for $Z^{\prime} \to n(\mu\mu)$ in C2, the score does not exhibit a dominant correlation with $n_{\text{leptons}}$. This suggests that the learned representation routes the signal into a highly specific leptonic region but the autoencoder's reconstruction error within that local neighborhood correlates more strongly with secondary kinematic features than with lepton multiplicity. The comparison between the local Wasserstein-distance rankings and the anomaly-score correlations reveals an important distinction between the physical observables that characterize the anomalous population and those associated with variations in the anomaly-score response. For the $Z'$ benchmark, lepton multiplicity provides the strongest separation from the local SM background, while the anomaly score shows stronger correlations with other kinematic observables, such as $H_T$ and $M_{jj}$. This shows that identifying the physical signature of an anomalous event population and understanding the behavior of the anomaly-scoring model are complementary diagnostic tasks, whose combined analysis provides information that cannot be inferred from global anomaly-detection performance alone.

% Requires \usepackage{subcaption} in the preamble.

\FloatBarrier
\section{Conclusion}
\label{sec:conclusion}

We studied the interplay between learned event representations, autoencoder-based AD, and latent-space diagnostics on \textsc{Collide-2V} under realistic HL-LHC conditions. Across the configurations and benchmark signals considered, representations learned with supervised objectives generally achieve higher linear-probe quality and anomaly-detection performance than their self-supervised counterparts. We also present a GMM-based pipeline that partitions the SM latent space, profiles each region through high-level observables without using labels, localizes the flagged events, and tests whether the anomaly score tracks the observables that separate them locally. This procedure recovers the defining physics of both processes (heavy-flavour content for $HH \to 4b$, lepton multiplicity for the $Z'$) and exposes a decoupling that aggregate metrics cannot capture. For example, the representation routes the $Z'$ signal into a well-defined leptonic region, but the local anomaly score is driven by secondary kinematics rather than lepton multiplicity. Requiring only embeddings, a per-event score, and a set of observables, this diagnostic procedure can be applied to any representation and anomaly-scoring model.

To address current limitations, our future work will focus on improving self-supervised representations. While process labels are available in simulated training samples, they are not observable on an event-by-event basis in collision data. Moreover, representations learned through supervision on simulated processes may inherit a dependence on the accuracy of the underlying physics and detector modeling. Developing effective self-supervised representations therefore remains important to reduce this dependence and to enable learning directly from experimental data, which will require overcoming the limitations of the self-supervised objectives considered here. To this end, our future works will therefore explore more advanced paradigms, such as JEPA or foundation model-like approaches, which are the natural candidates for learning robust representations directly from large-scale unlabeled data.

\section*{Acknowledgements}\label{Acknowledgement}
DG acknowledges support from Sapienza University of Rome (Avvio alla Ricerca 2025, Grant No. AR125199BE2808AE) and PNRR MUR project PE0000013-FAIR. We acknowledge ISCRA for awarding this project access to the LEONARDO supercomputer, owned by the EuroHPC Joint Undertaking, hosted by CINECA (Italy). TKA and PP are funded by the European Union (ERC, LEAP-40, grant agreement No. 101221186). Views and opinions expressed are however those of the author(s) only and do not necessarily reflect those of the European Union or the European Research Council Executive Agency. Neither the European Union nor the granting authority can be held responsible for them. This research was supported by COST Action CA24153 – Edge Deep Learning for Particle Physics (EPIGRAPHY), funded by the European Union.

\section*{Code Availability}
The code required to reproduce the results of this work is publicly available at
\url{https://github.com/DonatellaGenovese/foundation_model_testing_for_AD}. The
\textsc{Collide-2V} data set is publicly available from its authors~\cite{COLLIDE2V}.

\section*{Declaration on the use of AI tools}

The authors used large language model-based assistants to improve the
clarity and phrasing of the manuscript and to support routine coding
tasks. The authors reviewed and edited all text and code and take full
responsibility for the content of this work.

\begin{appendix}
\numberwithin{equation}{section}

\section{Experimental details}\label{app:experimental}

\subsection{Preprocessing and event representation}\label{sec:A1}

Dataset preprocessing proceeds in two stages: feature transformations are first applied to each variable, followed by normalization. Transverse momentum $p_T$, jet mass, and MET are mapped with $\log(1 + \cdot)$ to reduce dynamic range. Azimuthal  angles $\phi$ are encoded as $(\sin\phi,\cos\phi)$ to avoid discontinuities at the $\pm\pi$ boundary and are left un-normalized, as the encoding already lives in $[-1,1]$. Discrete variables are handled by kind: particle charge is one-hot encoded over $\{-1, 0, +1\}$, while the PUPPI jet b-tag, a bitmask over the six working points of the Delphes card, is kept as a single feature and min-max scaled. All remaining continuous variables ($\eta$, $E_\text{had}/E_\text{em}$, isolation, displacement $d_0$/$d_z$, $n_\text{neutral}$) undergo no transformation at this stage.

After transformation, each feature is independently normalized using robust normalization: median and interquartile range (IQR) are estimated from a subset of training files and used to standardize each feature as $(x - \text{median}) / \text{IQR}$. This is applied to all continuous features except $\phi$, the jet-multiplicity scalar, and one-hot encoded variables, which are left unchanged. The per-jet constituent counts $n_\text{neutral}$ is instead normalized like any other continuous feature. Normalization statistics are fitted once on the SM training data and applied identically to validation, test, and BSM signal sets. After preprocessing and normalization, each event is represented as a fixed-length vector obtained by zero-padding each object group to its maximum multiplicity (12 jets, 8 electrons, 8 muons, 8 photons). A binary mask distinguishes genuine zero values from padding and is used throughout augmentation to protect padded entries. For each object group the true multiplicity, before the top-k selection, is appended as a trailing scalar to the corresponding slice of the flat vector.

For models using the Transformer encoder, the preprocessed flat feature vector is converted into a token sequence. Each object group (jets, electrons, muons, photons, MET) contributes one token per object, formed by projecting its features to the common model dimension $d_\text{model}$ via a group-specific linear layer, for a maximum of 12 tokens for jets and 8 for electrons, muons, and photons. An additional event-level token is constructed by projecting the object-multiplicity counts appended to the flat vector during preprocessing. A learned type embedding is added to every token to identify its object group; no positional embedding is used, preserving permutation invariance within each group. For baseline experiments without learned representation, the full preprocessed flat vector (zero-padded to maximum multiplicity) is used directly as input to the autoencoder and probe linear classifier.

\begin{figure}[t]
  \centering
  \includegraphics[width=\textwidth]{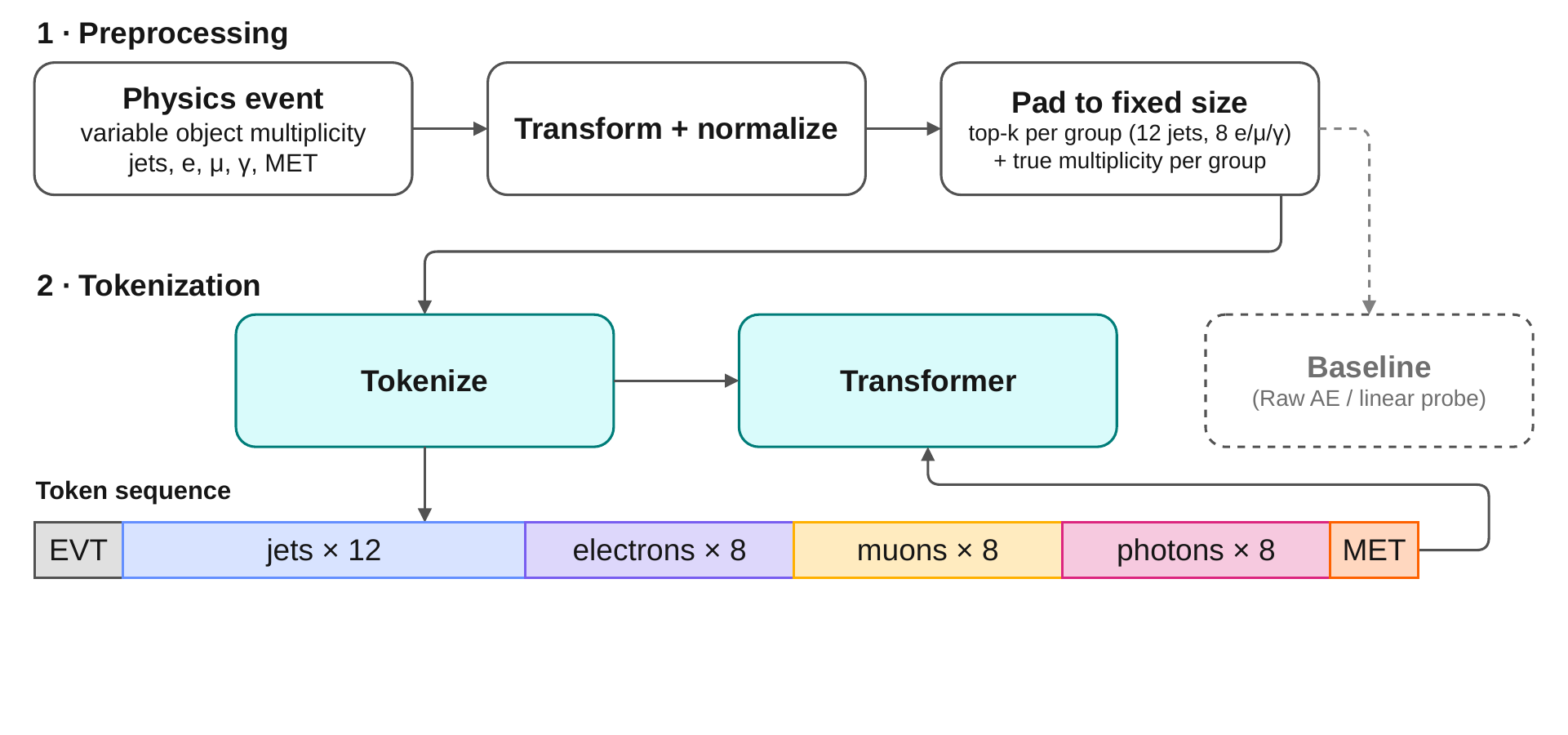}
  \caption{\textbf{Preprocessing and event representation.} Each event contains a variable number of reconstructed jets, electrons, muons, photons and MET. Features are transformed and normalized, then each group is truncated to its top-$k$ objects and zero padded to a fixed length, with the true per-group multiplicity appended as a trailing scalar. The resulting flat vector is tokenized by projecting every object slot to $d_\text{model}$, giving the token sequence shown below, color-coded by object group and preceded by an event token (EVT) built from the multiplicity counts, which the Transformer encoder then consumes. The dashed path denotes the baseline experiments, where the flat vector bypasses tokenization and is fed directly to the autoencoder or the linear probe.}
  \label{fig:preproc}
\end{figure}

\cref{fig:preproc} summarizes the complete pipeline, from the variable-length
list of reconstructed objects to the token sequence fed to the Transformer encoder,
together with the baseline path in which the flat vector is used  directly. The encoder is trained on 1{,}000{,}000 events per SM class (12{,}000{,}000 total), with
100{,}000 events per class held out for validation and 100{,}000 for testing.

\subsection{Augmentation operators and strategies}
\label{app:augmentation}
 
We adopt three physics-motivated augmentations following~\cite{Dillon2}, adapted to our normalized feature space, and add random masking as a fourth augmentation.
 
% -------------------------------------------------------------------
\paragraph{Azimuthal rotation.}
Following~\cite{Dillon2}, a single angle $\Delta\phi \sim \mathrm{Uniform}(0, 2\pi)$ is drawn independently for each view and applied simultaneously to all reconstructed objects. Since azimuthal angles are stored as $(\sin\phi, \cos\phi)$ pairs, the rotation is implemented as an exact $SO(2)$ transformation without approximation:
\begin{equation}
\begin{pmatrix} \sin\phi_i' \\ \cos\phi_i' \end{pmatrix}
=
\begin{pmatrix}
  \phantom{-}\cos\Delta\phi & \sin\Delta\phi \\
  -\sin\Delta\phi & \cos\Delta\phi
\end{pmatrix}
\begin{pmatrix} \sin\phi_i \\ \cos\phi_i \end{pmatrix},
\qquad
\Delta\phi \sim \mathrm{Uniform}(0,\, 2\pi),
\label{eq:aug_phi}
\end{equation}
for every object $i$ in the event. The same $\Delta\phi$ is shared across all objects, preserving all inter-object azimuthal separations $\Delta\phi_{ij}$. Padded objects with $(\sin\phi_i, \cos\phi_i) = (0,0)$ map identically to $(0,0)$ under Eq.~\eqref{eq:aug_phi}, so no masking is required for this augmentation.
 
% -------------------------------------------------------------------
\paragraph{Longitudinal boost.}
Hard-scattering processes at the LHC are approximately invariant under longitudinal Lorentz boosts, which shift all pseudorapidities by a common constant $\eta_i \mapsto \eta_i + \Delta\eta$. Inspired by the $\eta$-smearing of~\cite{Dillon2}, we implement this as a \emph{global} boost: a single displacement $\Delta\eta$ is drawn per event and added to the normalized pseudorapidity of every real object,
\begin{equation}
\tilde{\eta}_i' = \tilde{\eta}_i \;+\; m_i \cdot \frac{\Delta\eta}{\mathrm{IQR}_{\eta,g}},
\qquad
\Delta\eta \sim \mathcal{N}(0,\, \sigma_\eta^2),
\label{eq:aug_eta}
\end{equation}
where $\mathrm{IQR}_{\eta,g}$ is the interquartile range of the $\eta$ distribution for the object group $g$ (jets, electrons, muons or photons) to which object $i$ belongs, converting the physical displacement into the normalized feature space. The same $\Delta\eta$ is shared across all objects, preserving all inter-object $\Delta\eta_{ij}$ differences. We use $\sigma_\eta = 1.5$. Padded objects are protected by the mask $m_i$, defined in ~\cref{sec:A1}.
 
% -------------------------------------------------------------------
\paragraph{Transverse momentum smearing.}
Detector energy resolution introduces independent measurement uncertainties in the transverse momentum of each reconstructed object. Following the energy-smearing augmentation of~\cite{Dillon2}, the $p_\mathrm{T}$ of each real object is perturbed by independent Gaussian noise. In our pipeline the noise is applied in $\log(1+p_\mathrm{T})$ space, where $p_\mathrm{T}$ has been preprocessed, and then converted back to the normalized feature space:
\begin{equation}
\tilde{p}_{\mathrm{T},i}' = \tilde{p}_{\mathrm{T},i} \;+\; m_i \cdot
\frac{\varepsilon_i}{\mathrm{IQR}_{p_{\mathrm{T}},g}},
\qquad
\varepsilon_i \overset{\mathrm{i.i.d.}}{\sim} \mathcal{N}(0,\, \sigma_{p_\mathrm{T}}^2),
\label{eq:aug_pt}
\end{equation}
where $\mathrm{IQR}_{p_{\mathrm{T}},g}$ is the interquartile range of $\log(1+p_\mathrm{T})$ for object group $g$. Each object receives an independent noise draw $\varepsilon_i$, reflecting the fact that detector resolution effects are uncorrelated across particles. The same prescription is applied to event-level missing transverse energy (MET), using its own $\mathrm{IQR}_\mathrm{MET}$. We use $\sigma_{p_\mathrm{T}} = 0.1$. Padded objects are protected by the mask $m_i$, defined in ~\cref{sec:A1}.
 
% -------------------------------------------------------------------
\paragraph{Random masking.}
As an additional augmentation, we randomly zero parts of the event representation to simulate detector inefficiencies and missing particles. Two strategies are supported and selected as a hyperparameter.
 
\textit{Feature-level masking} independently zeroes each feature dimension $d$ with probability $p_\mathrm{mask}$:
\begin{equation}
\tilde{x}_d' = \tilde{x}_d \cdot b_d,
\qquad
b_d \overset{\mathrm{i.i.d.}}{\sim} \mathrm{Bernoulli}(1 - p_{\mathrm{mask}}).
\label{eq:aug_mask_feat}
\end{equation}
 
\textit{Particle-level masking} treats each reconstructed object as an atomic unit, zeroing all $F_g$ features of object $i$ jointly with a single Bernoulli draw:
\begin{equation}
\tilde{x}_{i,1:F_g}' = \tilde{x}_{i,1:F_g} \cdot b_i,
\qquad
b_i \overset{\mathrm{i.i.d.}}{\sim} \mathrm{Bernoulli}(1 - p_{\mathrm{mask}}),
\label{eq:aug_mask_part}
\end{equation}
where $F_g$ is the number of features per object in group $g$, and the scalar $b_i$ is broadcast across all $F_g$ features of object $i$. Scalar event-level quantities (MET) are treated as a single object and masked in their entirety under particle-level masking.
The masking probability $p_\text{mask}$ and the choice of strategy are treated as per model hyperparameters and reported in \cref{tab:hparams_final}.

\subsection{Ablation Study}
\label{app:ablation}
We conduct a systematic ablation study to identify the optimal configuration for each neural embedding model. All ablation experiments are performed on a reduced subset of the 12-class Standard Model dataset in order to keep the scan computationally tractable, and model quality is assessed via linear probe AUROC averaged over five independent random seeds. The ablation is structured in three stages: augmentation strategy, loss function hyperparameters and backbone architecture. Each stage fixes all other parameters at their default values and varies only the quantity under study. The remaining backbone and optimization hyperparameters, which are not varied in the ablation, are subsequently tuned with Optuna ~\cite{optuna} for each model; the resulting final configurations are reported in \cref{tab:hparams_final}.

\subsubsection{Augmentation Strategy}
\label{sec:ablation_augmentation}

The models SupCon, SimCLR and VICReg require an augmentation function to generate two views of the same event. We evaluate three strategies presented in \cref{app:augmentation}.

Results are reported in \cref{tab:ablation_aug}. For SupCon, all three strategies yield similar performance (AUROC range ${\sim}0.004$), with Physics augmentations achieving the best mean ($0.8991 \pm 0.0006$) and lowest variance across seeds but Random particle augmentation is competitive on average. 

For SimCLR, Random particle masking collapses performance to $0.7712 \pm 0.0003$ while Random feature masking achieves the best mean ($0.8514 \pm 0.0020$).

% VICReg results
For VICReg all three strategies yield ${\sim}0.72$, well below both SupCon and SimCLR. We select Physics augmentation, the most stable across seeds ($0.7203 \pm 0.0097$).

\begin{table}[htbp]
  \centering
  \caption{Linear probe macro-averaged AUROC (mean $\pm$ std over 5 seeds) for the three augmentation strategies.
           Bold denotes the best result per model.}
  \label{tab:ablation_aug}
  \begin{tabular}{lccc}
    \toprule
    \textbf{Augmentation} & \textbf{SupCon} & \textbf{SimCLR} & \textbf{VICReg} \\
    \midrule
    Random feature  & $0.8952 \pm 0.0009$ & $\mathbf{0.8514 \pm 0.0020}$ & $ \mathbf{0.7258 \pm 0.0300}$ \\
    Random particle & $0.8988 \pm 0.0009$ & $0.7712 \pm 0.0003$          & $0.7209 \pm 0.0235$ \\
    Physics         & $\mathbf{0.8991 \pm 0.0006}$ & $0.8491 \pm 0.0017$ & $0.7203 \pm 0.0097$ \\
    \bottomrule
  \end{tabular}
\end{table}

%─────────────────────────────────────────────────────────────────────────────
\subsubsection{Loss Function Hyperparameters}
\label{sec:ablation_loss}

\paragraph{VCReg and VICReg.}
VCReg augments the supervised cross-entropy objective of Eq.~\eqref{eq:vcreg} in the main text, with a variance term (weight $\alpha$) and a covariance term (weight $\beta$) applied to the intermediate representations; no invariance term is present, its role being subsumed by the supervised loss. We ablate each weight independently while holding the other at its default value ($\alpha=25$, $\beta=1$, following~\cite{VCReg}). Results are shown in \cref{tab:ablation_loss_vicreg}.

The variance weight $\alpha=25$ achieves the best mean AUROC ($0.9030 \pm 0.0007$). Reducing it to 5 or 10 degrades performance, while increasing it to 50 retains most of the gain but at the cost of higher seed variance. For the covariance weight, the value $\beta=1$ is optimal; larger values consistently reduce performance, suggesting that over-penalizing feature correlations is harmful for this dataset. The overall sensitivity to these hyperparameters is modest (total range ${\sim}0.003$), confirming that VCReg is stable once the variance collapse regularization is sufficiently strong.

For VICReg, we fix the variance term $\mu=25$ and the covariance term $\nu=1$ at their default values and ablate only the invariance term $\lambda \in \{0, 5, 10, 25, 50\}$ with physics augmentation fixed. Performance is flat across all values (${\sim}0.72$ AUROC), with $\lambda=25$ achieving marginally the best mean.

\begin{table}[htbp]
  \centering
  \caption{VCReg and VICReg loss ablation. Linear probe macro-averaged AUROC (mean $\pm$ std over 5 seeds).
           Left: VCReg variance weight scan with $\beta=1$ fixed.
           Center: VCReg covariance weight scan with $\alpha=25$ fixed.
           Right: VICReg invariance weight scan with $\mu=25$, $\nu=1$ fixed.
           Bold denotes the best result per column.}
  \label{tab:ablation_loss_vicreg}
  \begin{tabular}{ccc}
    \begin{tabular}[t]{cc}
      \toprule
      $\alpha$ & AUROC \\
      \midrule
       5  & $0.8998 \pm 0.0010$ \\
      10  & $0.9017 \pm 0.0003$ \\
      \textbf{25}  & $\mathbf{0.9030 \pm 0.0007}$ \\
      50  & $0.9020 \pm 0.0017$ \\
      \bottomrule
    \end{tabular}
    &
    \begin{tabular}[t]{cc}
      \toprule
      $\beta$ & AUROC \\
      \midrule
      \textbf{1}  & $\mathbf{0.9030 \pm 0.0007}$ \\
       5  & $0.9019 \pm 0.0012$ \\
      10  & $0.9007 \pm 0.0002$ \\
      \bottomrule
    \end{tabular}
    &
    \begin{tabular}[t]{cc}
      \toprule
      $\lambda$ & AUROC \\
      \midrule
       0  & $0.7192 \pm 0.0144$ \\
       5  & $0.7191 \pm 0.0318$ \\
      10  & $0.7239 \pm 0.0072$ \\
      \textbf{25} & $\mathbf{0.7269 \pm 0.0090}$ \\
      50  & $0.7170 \pm 0.0251$ \\
      \bottomrule
    \end{tabular}
  \end{tabular}
\end{table}

\paragraph{SupCon and SimCLR.}
For the contrastive models, the primary loss hyperparameter is the temperature $\tau$ of the supervised/self-supervised contrastive loss~\cite{SupCon}. For SupCon we additionally ablate the classification head weight $\lambda_\text{cls}$, which controls the contribution of a secondary cross-entropy term. We scan $\tau \in \{0.05, 0.07, 0.10, 0.15, 0.20\}$ and $\lambda_\text{cls} \in \{0.0, 0.05, 0.10, 0.20, 0.50\}$. Results are shown in Tables~\ref{tab:ablation_loss_temp} and~\ref{tab:ablation_loss_cw}. For SupCon, performance is nearly flat across all temperatures (range ${\sim}0.0002$), while $\lambda_\text{cls}$ shows a monotonic improvement with increasing values, with $\lambda_\text{cls}=0.50$ achieving the best result ($0.9019 \pm 0.0001$). For SimCLR, the best performance is obtained with $\tau=0.10$, which outperforms $\tau=0.20$ by $0.034$ AUROC, suggesting that harder negative mining is critical for the self-supervised objective.

% Tables
\begin{table}[htbp]
  \centering
  \caption{Temperature ablation for SupCon and SimCLR. Linear probe macro-averaged AUROC (mean $\pm$ std over 5 seeds). Bold denotes the best result per model.}
  \label{tab:ablation_loss_temp}
  \begin{tabular}{lcc}
    \toprule
    $\tau$ & \textbf{SupCon} & \textbf{SimCLR} \\
    \midrule
    0.05 & $0.9005 \pm 0.0006$ & $0.8670 \pm 0.0010$ \\
    0.07 & $0.9005 \pm 0.0004$ & $0.8643 \pm 0.0023$ \\
    0.10 & $\mathbf{0.9007 \pm 0.0005}$ & $\mathbf{0.8680 \pm 0.0015}$ \\
    0.15 & $0.9007 \pm 0.0004$ & $0.8405 \pm 0.0022$ \\
    0.20 & $0.9007 \pm 0.0002$ & $0.8343 \pm 0.0053$ \\
    \bottomrule
  \end{tabular}
\end{table}

\begin{table}[htbp]
  \centering
  \caption{Classification head weight ablation for SupCon ($\tau=0.10$ fixed). Linear probe macro-averaged AUROC (mean $\pm$ std over 5 seeds). Bold denotes the best result.}
  \label{tab:ablation_loss_cw}
  \begin{tabular}{cc}
    \toprule
    $\lambda_\text{cls}$ & \textbf{SupCon} \\
    \midrule
    0.00 & $0.8991 \pm 0.0006$ \\
    0.05 & $0.9000 \pm 0.0007$ \\
    0.10 & $0.9007 \pm 0.0005$ \\
    0.20 & $0.9010 \pm 0.0005$ \\
    0.50 & $\mathbf{0.9019 \pm 0.0001}$ \\
    \bottomrule
  \end{tabular}
\end{table}

\subsubsection{Backbone Architecture}
\label{sec:ablation_backbone}

We ablate the embedding dimension $d_{model}$; the remaining backbone and optimization hyperparameters are fixed via Optuna optimization and reported in \cref{tab:hparams_final}.

\cref{tab:ablation_dmodel} shows that all four objectives benefit from a larger embedding space. The sensitivity to $d_\text{model}$ is, however, strongly model-dependent: the supervised objectives vary only marginally across the whole scan, whereas the purely self-supervised ones improve considerably and also become appreciably more stable across seeds as $d_\text{model}$ grows. The label information thus appears to make representation quality largely insensitive to the width of the embedding, a capacity the self-supervised objectives instead need in order to avoid partial collapse.

\begin{table}[t]
    \centering
    \caption{%
        Linear probe macro-averaged AUROC for the $d_{\text{model}}$ ablation on the reduced ablation subset (\cref{app:ablation}). Values are mean $\pm$ standard deviation across 5 random seeds.
    }
    \label{tab:ablation_dmodel}
    \setlength{\tabcolsep}{6pt}
    \begin{tabular}{lrrrrr}
        \toprule
        \textbf{Model}
            & $d_{\text{model}}{=}32$
            & $d_{\text{model}}{=}64$
            & $d_{\text{model}}{=}128$
            & $d_{\text{model}}{=}256$
             \\
        \midrule
        VCReg
            & $0.898 \pm 0.001$
            & $0.902 \pm 0.001$
            & $\mathbf{0.903 \pm 0.001}$
            & $0.901 \pm 0.007$
             \\
        SupCon
            & $0.890 \pm 0.000$
            & $0.896 \pm 0.001$
            & $\mathbf{0.902 \pm 0.001}$
            & $0.902 \pm 0.001$
             \\
        SimCLR
            & $0.853 \pm 0.036$
            & $0.858 \pm 0.006$
            & $0.867 \pm 0.001$
            & $\mathbf{0.869 \pm 0.001}$
             \\
        VICReg
            & $0.686 \pm 0.030$ 
            & $0.717 \pm 0.004$ 
            & $0.721 \pm 0.024$ 
            & $\mathbf{0.739 \pm 0.015}$ 
             \\
        \bottomrule
    \end{tabular}
\end{table}

\subsection{Gaussian Mixture Model}
\label{app:gmm_ablation}

The interpretability pipeline of \cref{sec:adint} fits a Gaussian Mixture Model on the learned Standard Model embeddings. In order to do this we need to specify the space in which the mixture is fitted and the number of components $K$. 
In \cref{sec:ad_results} we report that the configuration that recovers stable AD performance results in terms of AD (for both proxy and real signal) is the VCReg with $d_{\text{model}}=256$. Since fitting a full-covariance mixture in $d_{\text{model}=256}$ is impractical, we first project to 64 dimensions with PCA.

To select $K$ we adopt an occupancy criterion: for a given $K$ we require every component to hold at least one fifth of the share it would hold under a uniform partition, that is, at least $1/(5K)$ of the events. The factor of one fifth is conventional.

What the criterion is meant to detect is a change in the nature of the partition. Beyond a certain $K$ the mixture stops reorganizing the space and starts splitting a tail off an existing component, and the resulting region reflects the number of components requested rather than any structure present in the data.

\begin{figure}[t]
  \centering
  \includegraphics[width=\textwidth]{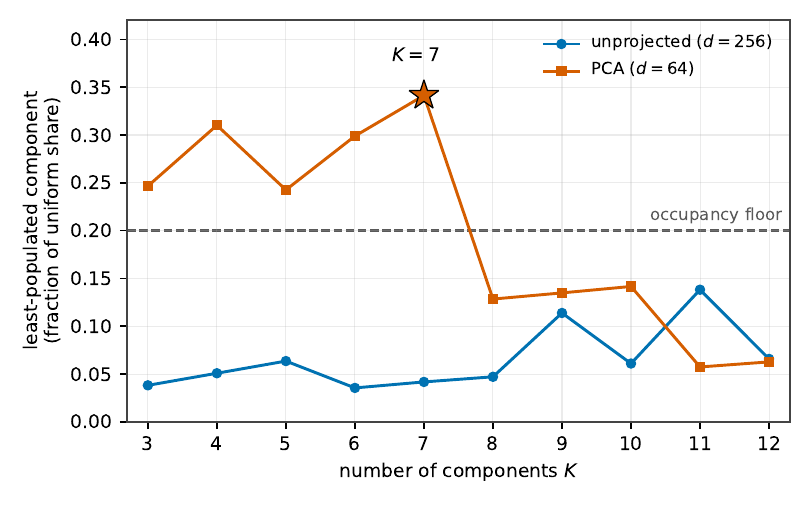}
  \caption{Occupancy of the least-populated mixture component as a function of $K$, as a fraction of the share it would hold under a uniform partition; the dashed line is the occupancy floor.}
  \label{fig:k_occupancy}
\end{figure}

The transition is sharp, against the fluctuation of the preceding
values, as \cref{fig:k_occupancy} shows: between $K = 7$ and $K = 8$ an eighth component appears holding $1.6\%$ of the events, drawn entirely from a single existing one: every event it contains was assigned to the same component at $K = 7$, while the other six retain between $91\%$ and $100\%$ of theirs. The least-populated component therefore drops from $34.1\%$ to $12.8\%$ of the uniform share. The requirement is satisfied for $K\le7$, and we take $K=7$, the finest such partition. 

%add a final table on GMM 

For the AD task we report the hyperparameters in \cref{tab:ae_hparams}.

\begin{table}[ht]
\centering
\caption{Final hyperparameter configuration per model, as used for the results of \cref{sec:results}.}
\label{tab:hparams_final}
\begin{tabular}{lccccc}
\toprule
\textbf{Parameter} & \textbf{SupCon} & \textbf{SimCLR} & \textbf{VCReg} & \textbf{CE}
& \textbf{VICReg} \\
\midrule
$d_\text{model}$                  & 128   & 256   & 128   & 256   & 256 \\
$n_\text{heads}$                  & 8     & 8     & 8     & 8     & 4 \\
$n_\text{layers}$                 & 6     & 8     & 4     & 4     & 4 \\
$d_\text{ff}$                     & 1024  & 2048  & 1024  & 2048  & 512 \\
Dropout                           & 0.10  & 0.226 & 0.10  & 0.10  & 0.10 \\
Projection dim.                   & 32    & 32    & ---   & ---   & 128 \\
Hidden projection dim.            & 128   & 256   & ---   & ---   & 256 \\
\midrule
Optimizer                         & AdamW & AdamW & AdamW & AdamW & AdamW \\
Learning rate                     & $5\times10^{-4}$ & $1.26\times10^{-4}$
                                  & $1.5\times10^{-4}$ & $2\times10^{-4}$ & $3\times10^{-4}$ \\
Weight decay                      & $10^{-5}$ & $3.58\times10^{-2}$
                                  & $5\times10^{-4}$ & $10^{-2}$ & $10^{-5}$ \\
Batch size                        & 512   & 1024  & 1024  & 1024  & 1024 \\
\midrule
Temperature $\tau$                & 0.10  & 0.10  & ---   & ---   & --- \\
Class.\ head $\lambda_\text{cls}$ & 0.50  & ---   & ---   & ---   & --- \\
Variance weight $\alpha$ ($\mu$)  & ---   & ---   & 25    & ---   & 25 \\
Covariance weight $\beta$ ($\nu$) & ---   & ---   & 1     & ---   & 1 \\
Invariance weight $\lambda$       & ---   & ---   & ---   & ---   & 25 \\
\midrule
Augmentation strategy             & Physics & Random feature & --- & --- & Physics \\
Mask prob.\ $p_\text{mask}$       & ---  & 0.448 & ---   & ---   & --- \\
\bottomrule
\end{tabular}
\end{table}

\begin{table}[ht]
\centering
\caption{Autoencoder hyperparameters, identical for the embedding-based and raw-feature experiments}
\label{tab:ae_hparams}
\begin{tabular}{lc}
\toprule
Parameter & Value \\
\midrule
Compression   & 4 \\
Depth         & 4 \\
Dropout       & 0.035 \\
Learning rate & $1.97\times10^{-4}$ \\
Weight decay  & $3.02\times10^{-5}$ \\
Train events  & 200,000 \\
\bottomrule
\end{tabular}
\end{table}

\subsection{Observable definitions}
\label{app:observables}

Observables are computed from the raw vectorized fields, before the transformations
and normalization of \cref{sec:A1}, so all quantities below are in their physical
units. Jets are those written by \textsc{Collide-2V} ($p_T > 15$~GeV, $|\eta| < 5$) and
electrons and muons are its \texttt{Electron} and \texttt{MuonTight} collections; no further kinematic or isolation requirement is applied. Objects are $p_T$-ordered in the dataset, so ``leading'' means highest $p_T$. The event representation keeps at most the twelve hardest jets and the eight hardest leptons of each flavour, and the observables below are computed on what it keeps.

\begin{itemize}
  \item $H_T = \sum_i p_T^{\,j_i}$ over all stored jets; $n_{\rm jets}$ their number.
  \item MET: magnitude of the \texttt{PUPPIMET} missing transverse momentum, the
        pileup-mitigated estimator appropriate at $\langle\mu\rangle = 200$.
  \item $n_{b\text{-tag}}$: number of jets passing the medium $b$-tagging working
        point, that is, with bit~1 of the \texttt{BTag} field set.
  \item $n_{\rm leptons}$: electrons plus tight muons, counted on the two collections
        above.
  \item $M_{jj}$, $|\Delta\eta_{jj}|$: invariant mass of the sum of the two hardest
        jets' four-momenta, built with $E = \sqrt{p_T^2\cosh^2\eta + m^2}$, and the
        absolute pseudorapidity gap between them.
  \item $M_T = \sqrt{2 p_T^{\ell}\,\mathrm{MET}\,(1 - \cos\Delta\phi)}$, using the
        highest-$p_T$ lepton of the combined electron and muon collection and its
        azimuthal separation from the MET direction.
\end{itemize}

$M_{jj}$ and $|\Delta\eta_{jj}|$ are defined only for events with at least two jets
and $M_T$ only for events with at least one lepton; events lacking the required
objects are excluded from the corresponding comparison rather than imputed. The
three observables are therefore evaluated on smaller populations than the other
five: $81\%$ of the Standard Model for $M_{jj}$ and $|\Delta\eta_{jj}|$, and $28\%$
for $M_T$. The global standard deviation that standardizes them in \cref{tab:wasserstein_rank} and \cref{fig:gmm_profiles} is estimated on the SM events of those same populations, so values from different rows are comparable in scale but are not measured on the same events.

The \textsc{Collide-2V} \texttt{BTag} field is a bitmask over the six working points defined
in the Delphes card. Measured against the generator-matched jet flavour, the bits
pair up into three levels: bits~0 and~3 give $b$-tagging efficiency and light-jet
mistag rate $0.82/0.119$, bits~1 and~4 give $0.65/0.018$, and bits~2 and~5 give
$0.47/0.0026$. We count jets with bit~1 set. The bits are independent draws rather
than nested cuts on a continuous discriminant, and jets pass a tighter level without passing a looser one, so testing the field as a whole is an OR over six trials that reaches a $0.24$ mistag rate, worse than any single level.

\newpage
\section{Additional results}\label{app:addresults} 

\subsection{Probe results}
\label{app:addresults}
\cref{tab:probe_auroc_perclass} breaks down the linear-probe AUROC of \cref{tab:probe_auroc} (main text) by individual SM process. The three supervised models are close throughout: VCReg has the highest AUROC in nine of the twelve classes, SupCon and CE in the remaining three. VICReg trails in every class, and SimCLR sits close to the raw feature baseline, exceeding it by a small margin in most classes and falling below it in $Z\to bb$ and $Z\to cc$. Performance is lowest for $W\to q\bar{q}$, $\gamma$, QCD inclusive and $Z\to qq$, whose
final states are the least distinctive of the set, and highest for the three $t\bar t$ channels, which all exceed $0.97$ for the supervised models. At fixed production mode, heavy-flavour final states are easier than light-flavour ones: $Z\to bb$ is separated better than $Z\to qq$ by every model, and $Z\to cc$ by every supervised model.

\begin{table}[htbp]
\centering
\caption{Per-class linear probe AUROC (mean $\pm$ std over 5 seeds) for each model in its final configuration. For CE the AUROC is obtained from the classifier's own test predictions.}
\label{tab:probe_auroc_perclass}
\resizebox{\textwidth}{!}{%
\begin{tabular}{l cccc cc}
\toprule
& \multicolumn{4}{c}{\textbf{Learned representation}} & \textbf{Sup.} & \textbf{Base} \\
\cmidrule(lr){2-5}\cmidrule(lr){6-6}\cmidrule(lr){7-7}
\textbf{Class}
& SupCon & SimCLR & VCReg & VICReg & CE & Raw features \\
& \footnotesize$d{=}128$ & \footnotesize$d{=}256$ & \footnotesize$d{=}128$
& \footnotesize$d{=}256$ & \footnotesize$d{=}256$ & \\
\midrule
QCD incl.\ ($H_\mathrm{T}{>}50$\,GeV) & $0.8478 \pm 0.0056$ & $0.8143 \pm 0.0028$ & $\mathbf{0.8517 \pm 0.0005}$ & $0.6765 \pm 0.0521$ & $0.8443 \pm 0.0003$ & $0.8067 \pm 0.0018$ \\
QCD $\to b\bar{b}$                     & $0.8779 \pm 0.0016$ & $0.8107 \pm 0.0019$ & $\mathbf{0.8786 \pm 0.0008}$ & $0.6710 \pm 0.0420$ & $0.8684 \pm 0.0007$ & $0.8091 \pm 0.0020$ \\
$W \to \ell\nu$                        & $0.9483 \pm 0.0012$ & $0.9219 \pm 0.0016$ & $\mathbf{0.9486 \pm 0.0001}$ & $0.8187 \pm 0.0361$ & $0.9372 \pm 0.0002$ & $0.9219 \pm 0.0019$ \\
$W \to q\bar{q}$                       & $0.8291 \pm 0.0044$ & $0.7957 \pm 0.0019$ & $\mathbf{0.8333 \pm 0.0005}$ & $0.7056 \pm 0.0264$ & $0.8243 \pm 0.0006$ & $0.7909 \pm 0.0015$ \\
$Z \to \nu\nu$+jets                    & $0.9174 \pm 0.0013$ & $0.8947 \pm 0.0020$ & $\mathbf{0.9185 \pm 0.0002}$ & $0.7451 \pm 0.0323$ & $0.9103 \pm 0.0003$ & $0.8889 \pm 0.0017$ \\
$Z \to qq$ (uds)                       & $0.8568 \pm 0.0015$ & $0.8144 \pm 0.0013$ & $\mathbf{0.8583 \pm 0.0005}$ & $0.7067 \pm 0.0327$ & $0.8530 \pm 0.0005$ & $0.8135 \pm 0.0027$ \\
$Z \to bb$                             & $0.9227 \pm 0.0036$ & $0.8708 \pm 0.0014$ & $\mathbf{0.9310 \pm 0.0005}$ & $0.7545 \pm 0.0535$ & $0.9191 \pm 0.0006$ & $0.8727 \pm 0.0013$ \\
$Z \to cc$                             & $0.8675 \pm 0.0032$ & $0.7847 \pm 0.0014$ & $\mathbf{0.8735 \pm 0.0005}$ & $0.6904 \pm 0.0268$ & $0.8616 \pm 0.0013$ & $0.7853 \pm 0.0019$ \\
$\gamma$                         & $0.8459 \pm 0.0032$ & $0.8270 \pm 0.0015$ & $0.8498 \pm 0.0005$ & $0.7512 \pm 0.0182$ & $\mathbf{0.8613 \pm 0.0003}$ & $0.8177 \pm 0.0021$ \\
$t\bar{t}$ (all-hadronic)                  & $\mathbf{0.9888 \pm 0.0004}$ & $0.9767 \pm 0.0013$ & $0.9886 \pm 0.0001$ & $0.9269 \pm 0.0190$ & $\mathbf{0.9888 \pm 0.0001}$ & $0.9711 \pm 0.0004$ \\
$t\bar{t}$ (semi-leptonic)                 & $\mathbf{0.9732 \pm 0.0011}$ & $0.9425 \pm 0.0021$ & $0.9726 \pm 0.0004$ & $0.8799 \pm 0.0164$ & $0.9723 \pm 0.0002$ & $0.9399 \pm 0.0006$ \\
$t\bar{t}$ (all-leptonic)                  & $0.9857 \pm 0.0008$ & $0.9674 \pm 0.0012$ & $\mathbf{0.9858 \pm 0.0001}$ & $0.8917 \pm 0.0327$ & $0.9855 \pm 0.0002$ & $0.9666 \pm 0.0005$ \\
\midrule
Macro-averaged AUROC                             & $0.9051 \pm 0.0020$ & $0.8684 \pm 0.0014$ & $\mathbf{0.9075 \pm 0.0002}$ & $0.7682 \pm 0.0314$ & $0.9022 \pm 0.0004$ & $0.8653 \pm 0.0007$ \\
\bottomrule
\end{tabular}%
}
\end{table}

\subsection{Sensitivity across held-out processes}
\label{app:signals}

\cref{tab:heldout} extends the evaluation of \cref{sec:ad_results} to three further held-out proxies, VBF $H\to b\bar b$, $ggH\to\tau\tau$ and $HH\to b\bar b\tau\tau$, chosen to span a range of kinematic distance from the QCD background on which the autoencoder is trained. \cref{tab:heldout_bsm} reports two additional BSM signals, $H\to aa\to 4b$ and $H\to aa\to 4\tau$, covering topologies the proxies do not reach, together with $Z^{\prime} \to n(\mu\mu)$ for reference.

The two supervised objectives lead on every process. SupCon and VCReg are the best two models on all three proxies, and on the two BSM signals both exceed the Raw AE baseline by a wide margin: on $H\to aa\to 4b$ VCReg gains a factor $3.7$ in signal efficiency. Their advantage over the baseline is largest on the soft, low-$H_\mathrm{T}$ signals and smallest where the signature is explicit, as on $Z^{\prime} \to n(\mu\mu)$, where every model and the raw baseline exceed $0.97$ AUROC except VICReg.

The two supervised objectives differ in where they peak: on the proxies both improve with $d_\text{model}$ up to $256$, whereas on the BSM signals the dependence is less regular: VCReg peaks at $d_\text{model}=128$ on both, SupCon at $d_\text{model}=32$ on $H\to aa\to 4\tau$, while on $H\to aa\to 4b$ SupCon still improves up to $d_\text{model}=256$. Both fall below the Raw AE baseline at $d_\text{model}=32$ on VBF $H\to b\bar b$, the process on which the baseline is strongest relative to the learned representations.

The self-supervised objectives remain below the baseline on all proxies at every dimension. On the BSM signals, SimCLR is marginally above the baseline on $Z^{\prime} \to n(\mu\mu)$ and exceeds it in AUROC on $H\to aa\to 4\tau$ at $d_{\text{model}} = 64$, while being compatible with it at 32 and 128. SimCLR is worst at $d_{\text{model}} = 128$ on every process except $H \to aa \to 4\tau$, where $d_{\text{model}} = 32$ and 128 are compatible within their dispersions. VICReg carries the largest seed-to-seed dispersions of all models, reaching AUROC below $0.5$ on $Z^{\prime}$ at $d_\text{model}=128$. 
The signal efficiencies of \Cref{tab:heldout} and \Cref{tab:heldout_bsm} are quoted at the threshold that accepts $10\%$ of QCD events. \Cref{tab:sm_acceptance} reports, for VCReg with $d_{\text{model}} = 256$, the acceptance of the remaining eleven Standard Model processes at that same threshold. Processes whose final state resembles QCD pass at close to the nominal rate: $\gamma$, $Z \to qq$ and $W \to q\bar{q}$ at $10-13\%$. Those carrying leptons, missing transverse momentum or heavy-flavour jets pass far more often: $71\%$ of $W \to \ell\nu$, $64\%$ of $Z \to b\bar{b}$ and more than $90\%$ of each $t\bar{t}$ channel.

\begin{table}[htbp]
\centering
\scriptsize
\setlength{\tabcolsep}{3pt}
\caption{Held-out performance on four processes absent from training: AUROC and
signal efficiency at 10\% background acceptance (TPR @ FPR = 10\%), mean $\pm$ std
over 5 seeds. Raw AE is trained directly on preprocessed kinematic features. Bold
marks the best entry in each column; differences smaller than the quoted
uncertainties are not significant.}
\label{tab:heldout}
\resizebox{\textwidth}{!}{%
\begin{tabular}{ll cc cc cc cc}
\toprule
& & \multicolumn{2}{c}{\boldmath$HH\to4b$}
  & \multicolumn{2}{c}{\textbf{VBF} $H\to b\bar{b}$}
  & \multicolumn{2}{c}{$ggH\to\tau\tau$}
  & \multicolumn{2}{c}{$HH\to b\bar{b}\tau\tau$} \\
\cmidrule(lr){3-4}\cmidrule(lr){5-6}\cmidrule(lr){7-8}\cmidrule(lr){9-10}
\textbf{Model} & $d_\text{model}$
 & AUROC & TPR\,[\%] & AUROC & TPR\,[\%] & AUROC & TPR\,[\%] & AUROC & TPR\,[\%] \\
\midrule
Raw AE & --- & $0.931 \pm 0.001$ & $77.3 \pm 0.7$ & $0.863 \pm 0.002$ & $51.0 \pm 1.1$ & $0.815 \pm 0.003$ & $46.5 \pm 0.7$ & $0.939 \pm 0.001$ & $79.8 \pm 0.6$ \\
\midrule
\multirow{4}{*}{SupCon}
 & 32  & $0.892 \pm 0.031$ & $65.2 \pm 9.6$ & $0.753 \pm 0.014$ & $38.8 \pm 2.3$ & $0.889 \pm 0.021$ & $74.5 \pm 3.7$ & $0.967 \pm 0.003$ & $91.5 \pm 0.4$ \\
 & 64  & $0.937 \pm 0.017$ & $79.9 \pm 6.8$ & $0.851 \pm 0.029$ & $54.4 \pm 7.8$ & $0.899 \pm 0.006$ & $\mathbf{76.0 \pm 1.5}$ & $0.977 \pm 0.005$ & $93.7 \pm 1.8$ \\
 & 128 & $0.943 \pm 0.005$ & $82.4 \pm 1.9$ & $0.878 \pm 0.026$ & $61.5 \pm 5.8$ & $0.886 \pm 0.008$ & $71.6 \pm 1.5$ & $0.979 \pm 0.003$ & $94.2 \pm 1.0$ \\
 & 256 & $0.954 \pm 0.009$ & $86.4 \pm 4.2$ & $0.906 \pm 0.011$ & $68.6 \pm 4.4$ & $\mathbf{0.901 \pm 0.004}$ & $74.3 \pm 0.9$ & $\mathbf{0.985 \pm 0.002}$ & $95.8 \pm 0.7$ \\
\midrule
\multirow{4}{*}{SimCLR}
 & 32  & $0.663 \pm 0.057$ & $37.5 \pm 6.9$ & $0.540 \pm 0.050$ & $21.0 \pm 5.1$ & $0.621 \pm 0.025$ & $29.0 \pm 4.0$ & $0.687 \pm 0.049$ & $42.0 \pm 6.5$ \\
 & 64  & $0.553 \pm 0.057$ & $24.1 \pm 5.9$ & $0.454 \pm 0.033$ & $15.6 \pm 3.8$ & $0.584 \pm 0.021$ & $29.2 \pm 4.4$ & $0.602 \pm 0.046$ & $31.0 \pm 6.5$ \\
 & 128 & $0.502 \pm 0.069$ & $19.7 \pm 8.4$ & $0.406 \pm 0.044$ & $10.1 \pm 4.8$ & $0.566 \pm 0.035$ & $25.8 \pm 6.5$ & $0.571 \pm 0.065$ & $29.2 \pm 10.6$ \\
 & 256 & $0.756 \pm 0.047$ & $44.5 \pm 7.5$ & $0.603 \pm 0.048$ & $23.1 \pm 5.6$ & $0.685 \pm 0.025$ & $38.5 \pm 4.0$ & $0.775 \pm 0.036$ & $51.9 \pm 6.0$ \\
\midrule
\multirow{4}{*}{VCReg}
 & 32  & $0.937 \pm 0.025$ & $78.9 \pm 9.6$ & $0.824 \pm 0.022$ & $37.4 \pm 8.7$ & $0.827 \pm 0.018$ & $53.8 \pm 6.3$ & $0.899 \pm 0.023$ & $62.9 \pm 10.9$ \\
 & 64  & $0.937 \pm 0.012$ & $80.2 \pm 4.4$ & $0.830 \pm 0.021$ & $34.1 \pm 4.8$ & $0.825 \pm 0.024$ & $52.7 \pm 9.0$ & $0.929 \pm 0.011$ & $75.8 \pm 5.4$ \\
 & 128 & $0.959 \pm 0.005$ & $90.0 \pm 2.5$ & $0.890 \pm 0.009$ & $60.3 \pm 2.9$ & $0.866 \pm 0.010$ & $63.3 \pm 2.9$ & $0.967 \pm 0.004$ & $91.9 \pm 1.6$ \\
 & 256 & $\mathbf{0.982 \pm 0.004}$ & $\mathbf{97.2 \pm 1.7}$ & $\mathbf{0.924 \pm 0.017}$ & $\mathbf{76.6 \pm 6.9}$ & $0.882 \pm 0.009$ & $67.7 \pm 2.2$ & $0.983 \pm 0.002$ & $\mathbf{96.6 \pm 0.9}$ \\
\midrule
\multirow{4}{*}{VICReg}
 & 32  & $0.853 \pm 0.022$ & $63.0 \pm 5.8$ & $0.744 \pm 0.033$ & $37.0 \pm 5.8$ & $0.701 \pm 0.035$ & $29.1 \pm 9.9$ & $0.846 \pm 0.016$ & $58.3 \pm 10.3$ \\
 & 64  & $0.846 \pm 0.046$ & $62.3 \pm 3.9$ & $0.722 \pm 0.074$ & $37.0 \pm 3.6$ & $0.669 \pm 0.078$ & $31.7 \pm 12.1$ & $0.832 \pm 0.055$ & $60.3 \pm 8.4$ \\
 & 128 & $0.796 \pm 0.076$ & $52.7 \pm 21.8$ & $0.673 \pm 0.046$ & $30.1 \pm 10.4$ & $0.643 \pm 0.058$ & $28.2 \pm 7.3$ & $0.799 \pm 0.020$ & $53.7 \pm 8.3$ \\
 & 256 & $0.785 \pm 0.059$ & $53.4 \pm 18.7$ & $0.675 \pm 0.019$ & $32.4 \pm 8.5$ & $0.654 \pm 0.070$ & $29.1 \pm 5.1$ & $0.793 \pm 0.030$ & $53.9 \pm 10.2$ \\
\bottomrule
\end{tabular}%
}
\end{table}

\begin{table}[t]
\centering
\caption{AUROC and signal efficiency at 10\% background acceptance (mean $\pm$ std over 5 seeds). For each process, 20{,}000 events are drawn at random from a dedicated production of 50{,}000, matching the test statistics of the proxy signals of Table~\ref{tab:heldout}; the QCD reference is the same as for $HH \to 4b$.}
\label{tab:heldout_bsm}
\resizebox{\textwidth}{!}{%
\begin{tabular}{llcccccc}
\toprule
 & & \multicolumn{2}{c}{$H \to aa \to 4b$} & \multicolumn{2}{c}{$H \to aa \to 4\tau$} & \multicolumn{2}{c}{$Z' \to n(\mu\mu)$} \\
\cmidrule(lr){3-4}\cmidrule(lr){5-6}\cmidrule(lr){7-8}
Model & $d_{\text{model}}$ & AUROC & TPR@10\% & AUROC & TPR@10\% & AUROC & TPR@10\% \\
\midrule
Raw AE & --- & $0.640 \pm 0.001$ & $18.5 \pm 0.3$ & $0.681 \pm 0.004$ & $25.1 \pm 0.3$ & $0.984 \pm 0.001$ & $96.8 \pm 0.1$ \\
\midrule
\multirow{4}{*}{SupCon} & 32 & $0.739 \pm 0.041$ & $38.6 \pm 5.7$ & $\mathbf{0.876 \pm 0.014}$ & $\mathbf{70.8 \pm 2.9}$ & $\mathbf{0.996 \pm 0.001}$ & $\mathbf{99.0 \pm 0.2}$ \\
 & 64 & $0.758 \pm 0.018$ & $39.8 \pm 3.3$ & $0.872 \pm 0.009$ & $70.5 \pm 2.1$ & $0.995 \pm 0.001$ & $99.0 \pm 0.1$ \\
 & 128 & $0.777 \pm 0.024$ & $44.0 \pm 4.9$ & $0.842 \pm 0.011$ & $62.8 \pm 1.9$ & $0.995 \pm 0.001$ & $98.9 \pm 0.1$ \\
 & 256 & $0.829 \pm 0.014$ & $52.8 \pm 3.7$ & $0.854 \pm 0.009$ & $65.3 \pm 1.6$ & $0.995 \pm 0.001$ & $98.9 \pm 0.1$ \\
\midrule
\multirow{4}{*}{SimCLR} & 32 & $0.582 \pm 0.009$ & $18.1 \pm 1.6$ & $0.683 \pm 0.008$ & $32.1 \pm 2.1$ & $0.992 \pm 0.002$ & $98.0 \pm 0.5$ \\
 & 64 & $0.591 \pm 0.016$ & $23.1 \pm 3.5$ & $0.725 \pm 0.016$ & $42.0 \pm 3.4$ & $0.996 \pm 0.001$ & $98.9 \pm 0.3$ \\
 & 128 & $0.576 \pm 0.017$ & $16.4 \pm 2.8$ & $0.690 \pm 0.018$ & $35.6 \pm 4.4$ & $0.990 \pm 0.002$ & $97.5 \pm 0.5$ \\
 & 256 & $0.607 \pm 0.016$ & $21.5 \pm 3.2$ & $0.717 \pm 0.014$ & $41.7 \pm 3.7$ & $0.994 \pm 0.001$ & $98.6 \pm 0.4$ \\
\midrule
\multirow{4}{*}{VCReg} & 32 & $0.857 \pm 0.011$ & $60.3 \pm 4.0$ & $0.839 \pm 0.010$ & $57.9 \pm 3.2$ & $0.977 \pm 0.015$ & $95.4 \pm 4.4$ \\
 & 64 & $0.850 \pm 0.030$ & $55.6 \pm 9.4$ & $0.850 \pm 0.018$ & $61.2 \pm 4.6$ & $0.991 \pm 0.002$ & $98.4 \pm 0.4$ \\
 & 128 & $\mathbf{0.868 \pm 0.018}$ & $\mathbf{68.8 \pm 8.1}$ & $0.851 \pm 0.012$ & $63.3 \pm 3.5$ & $0.991 \pm 0.004$ & $98.4 \pm 0.5$ \\
 & 256 & $0.838 \pm 0.033$ & $59.1 \pm 11.8$ & $0.818 \pm 0.014$ & $58.1 \pm 4.1$ & $0.991 \pm 0.001$ & $98.4 \pm 0.2$ \\
\midrule
\multirow{4}{*}{VICReg} & 32 & $0.539 \pm 0.036$ & $12.2 \pm 10.0$ & $0.543 \pm 0.109$ & $14.0 \pm 17.1$ & $0.618 \pm 0.218$ & $25.8 \pm 40.9$ \\
 & 64 & $0.536 \pm 0.039$ & $11.7 \pm 5.9$ & $0.571 \pm 0.112$ & $18.4 \pm 16.2$ & $0.629 \pm 0.236$ & $27.0 \pm 30.6$ \\
 & 128 & $0.500 \pm 0.029$ & $7.2 \pm 0.7$ & $0.541 \pm 0.093$ & $15.1 \pm 13.1$ & $0.492 \pm 0.043$ & $7.3 \pm 1.9$ \\
 & 256 & $0.533 \pm 0.028$ & $8.3 \pm 2.3$ & $0.572 \pm 0.057$ & $13.8 \pm 9.1$ & $0.530 \pm 0.065$ & $8.3 \pm 2.1$ \\
\bottomrule
\end{tabular}%
}
\end{table}

\begin{table}[t]
\centering
\caption{AUROC against QCD and acceptance of each Standard Model process at the autoencoder threshold that accepts 10\% of QCD events, for VCReg with $d_{\text{model}} = 256$. The threshold is calibrated on QCD validation events for each encoder seed and applied unchanged to 20,000 test events per process; mean $\pm$ std over 5 seeds.}
\label{tab:sm_acceptance}
\begin{tabular}{lcc}
\toprule
\textbf{SM process} & \textbf{AUROC} & \textbf{Acceptance [\%]} \\
\midrule
QCD incl.\ ($H_\mathrm{T}{>}50$\,GeV) & --- & $9.6 \pm 0.3$ \\
$\gamma$ & $0.436 \pm 0.015$ & $10.0 \pm 0.6$ \\
$Z \to qq$ ($uds$) & $0.543 \pm 0.015$ & $11.1 \pm 0.8$ \\
$W \to q\bar{q}$ & $0.558 \pm 0.017$ & $13.1 \pm 1.3$ \\
$Z \to cc$ & $0.686 \pm 0.014$ & $28.2 \pm 3.0$ \\
$Z \to \nu\nu$ + jets & $0.676 \pm 0.023$ & $36.2 \pm 2.4$ \\
QCD $\to b\bar{b}$ & $0.719 \pm 0.025$ & $38.8 \pm 7.4$ \\
$Z \to bb$ & $0.834 \pm 0.015$ & $64.2 \pm 4.9$ \\
$W \to \ell\nu$ & $0.850 \pm 0.012$ & $70.7 \pm 0.9$ \\
$t\bar{t}$ (all-hadronic) & $0.970 \pm 0.005$ & $92.9 \pm 2.2$ \\
$t\bar{t}$ (semi-leptonic) & $0.991 \pm 0.001$ & $98.4 \pm 0.5$ \\
$t\bar{t}$ (all-leptonic) & $0.997 \pm 0.000$ & $99.5 \pm 0.1$ \\
\bottomrule
\end{tabular}
\end{table}

\subsection{Additional interpretability results}
\label{app:xai_extra}

This appendix collects the material underlying the latent space diagnostics of \cref{sec:results_gmm}.

\begin{figure}[p]
\centering
\begin{subfigure}[b]{\textwidth}
    \centering
    \includegraphics[width=\textwidth]{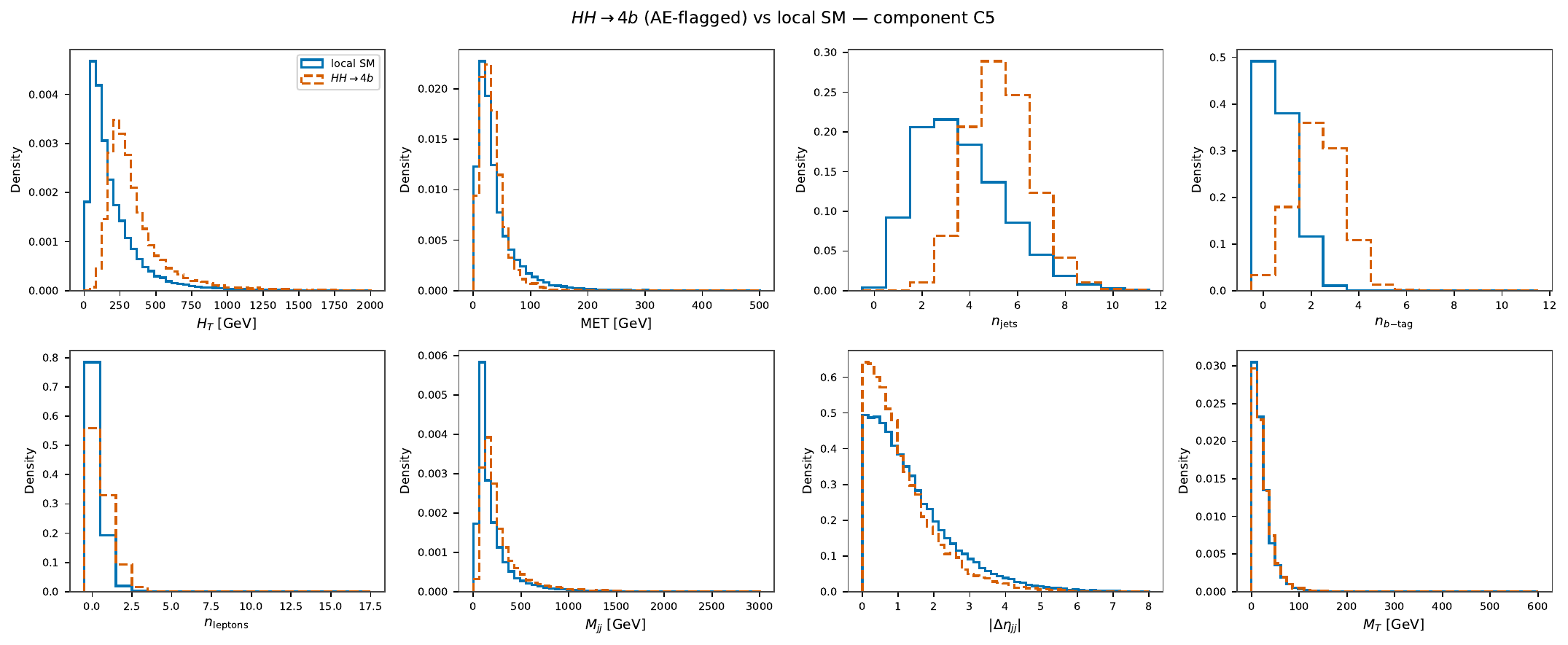}
    \caption{C5, holding $55.6\%$ of the flagged signal.}
\end{subfigure}

\vspace{0.8em}

\begin{subfigure}[b]{\textwidth}
    \centering
    \includegraphics[width=\textwidth]{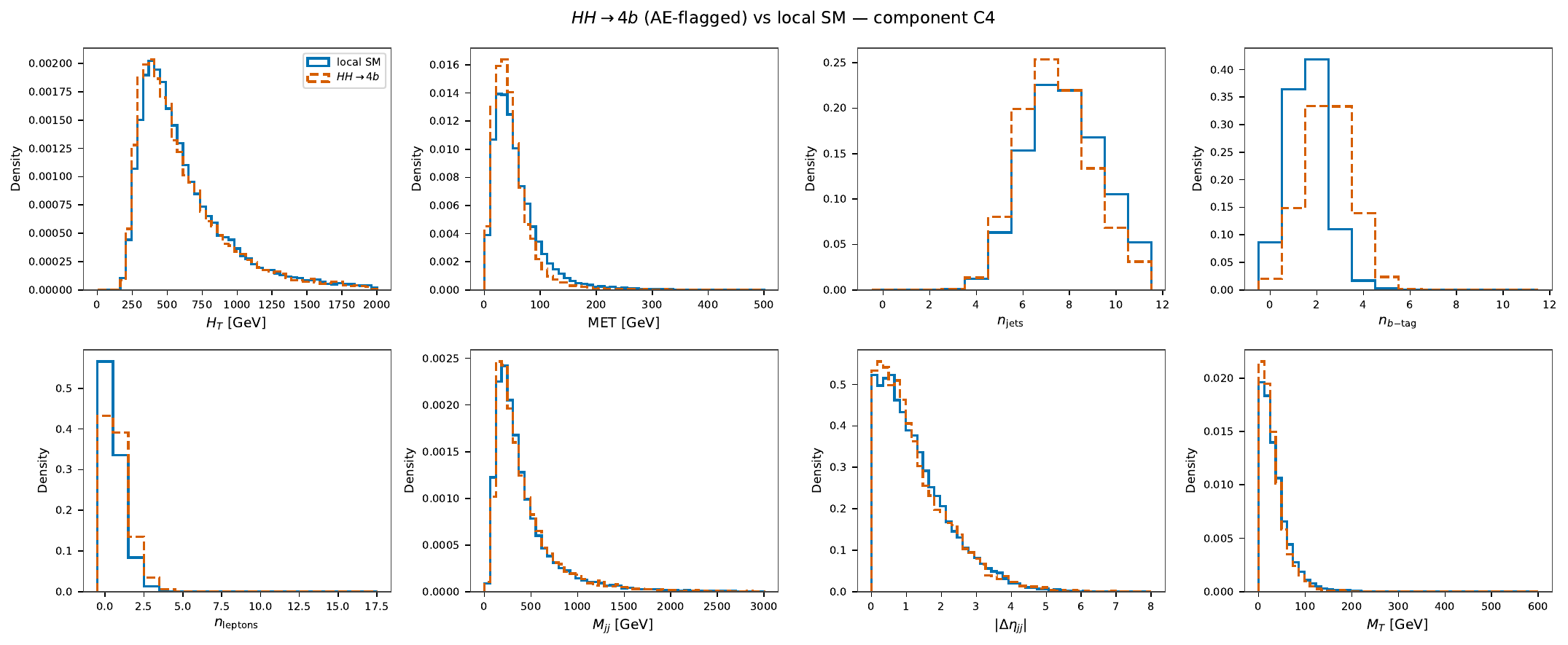}
    \caption{C4, holding $43.8\%$ of the flagged signal.}
\end{subfigure}

\caption{Distributions of the eight high-level observables for the AE-flagged $HH \to 4b$ signal (dashed) and the local Standard Model background (solid), shown separately for each mixture component that holds at least $5\%$ of the flagged signal. Every
histogram is normalized to unit area.}
\label{fig:dist8_hh4b}
\end{figure}

\begin{figure}[p]
\centering
\begin{subfigure}[b]{\textwidth}
    \centering
    \includegraphics[width=\textwidth]{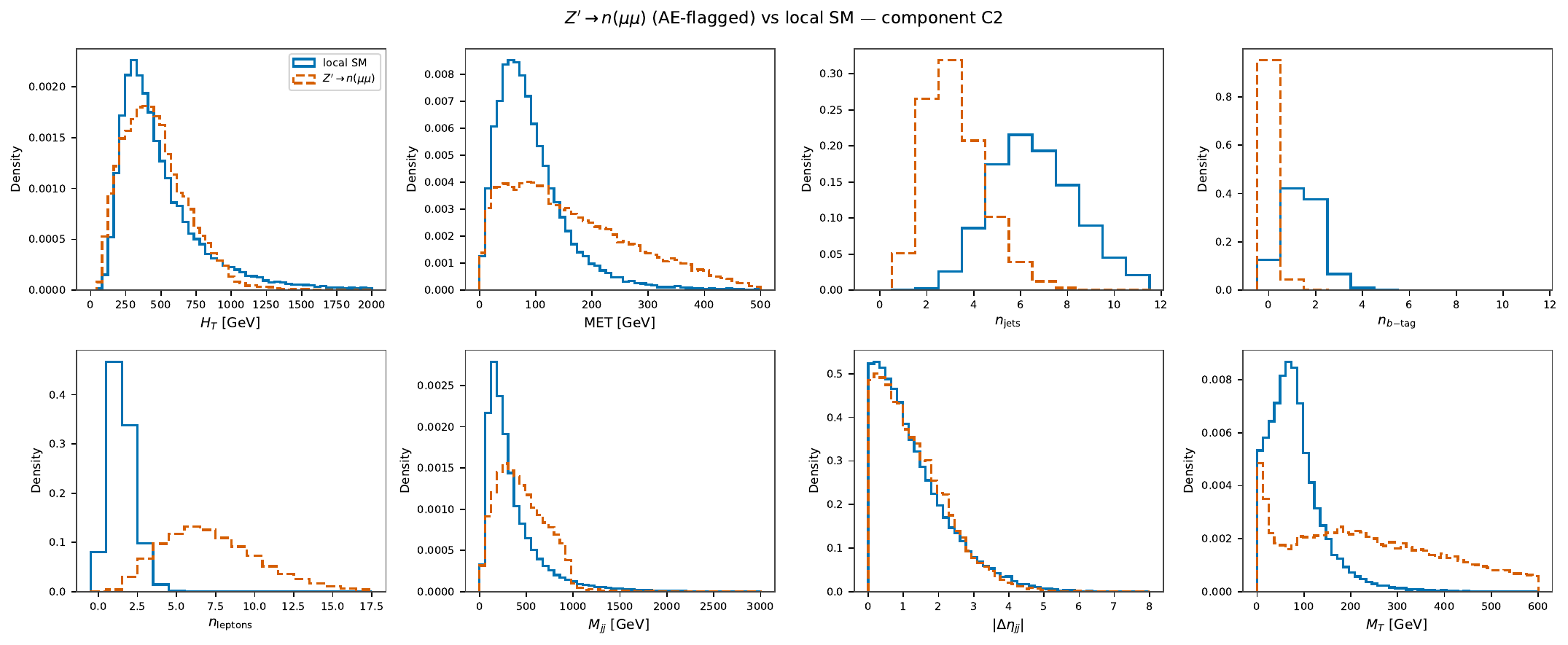}
    \caption{C2, holding $88\%$ of the flagged signal.}
\end{subfigure}

\vspace{0.8em}

\begin{subfigure}[b]{\textwidth}
    \centering
    \includegraphics[width=\textwidth]{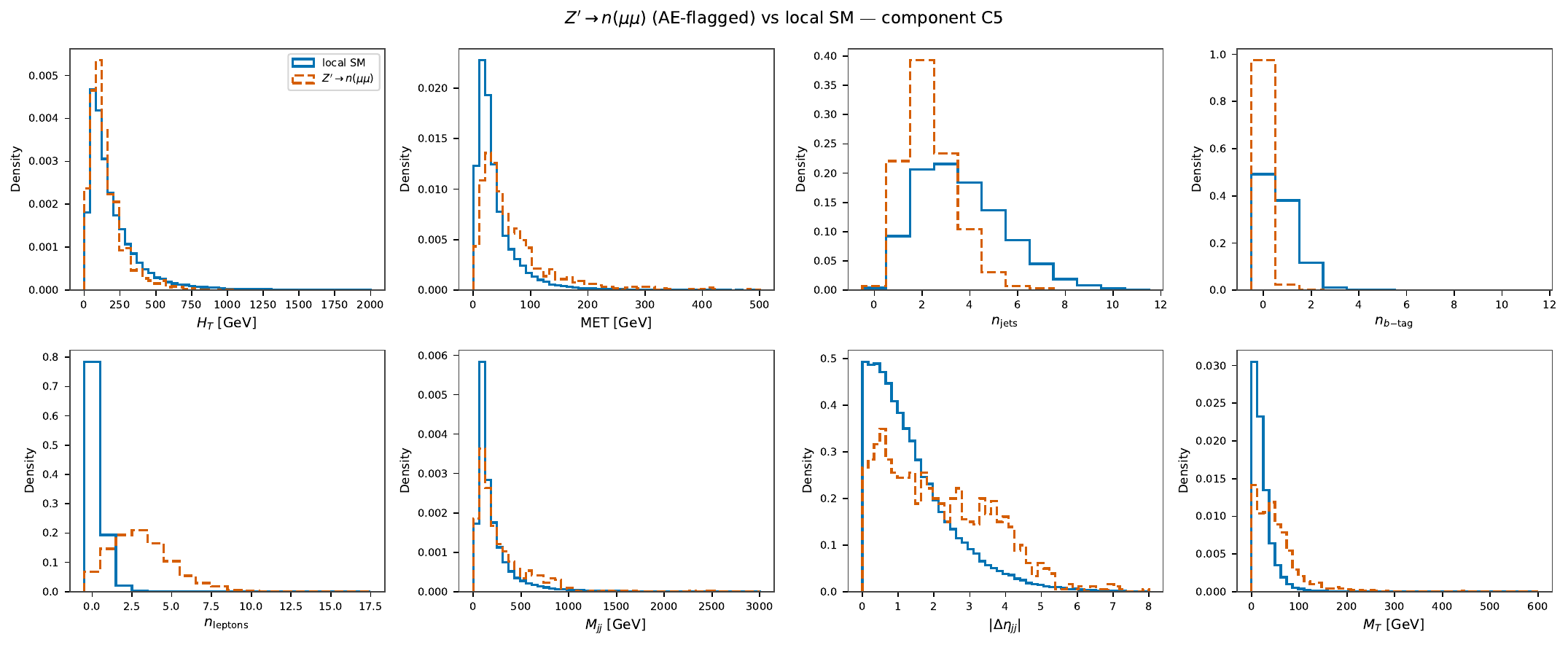}
    \caption{C5, holding $7.2\%$ of the flagged signal.}
\end{subfigure}

\caption{Distributions of the eight high-level observables for the AE-flagged $Z^{\prime} \to n(\mu\mu)$ signal (dashed) and the local Standard Model background (solid), shown separately for each mixture component that holds at least $5\%$ of the flagged signal. Every histogram is normalized to unit area.}

\label{fig:dist8_hvdilep}
\end{figure}

Figures~\ref{fig:dist8_hh4b} and \ref{fig:dist8_hvdilep} report the full 1D distributions of all eight high-level observables for the AE-flagged signals against the local SM background, in the components where each signal is localized. These are the distributions behind the Wasserstein ranking of \cref{tab:wasserstein_rank}, of which the main text shows only the leading observable. In both cases the top-ranked variable carries a visible separation while the remaining observables overlap far more closely, and the gap is markedly wider for $Z^{\prime} \to n(\mu\mu)$ than for the proxy.

\begin{figure}[t]
\centering
\includegraphics[width=\textwidth]{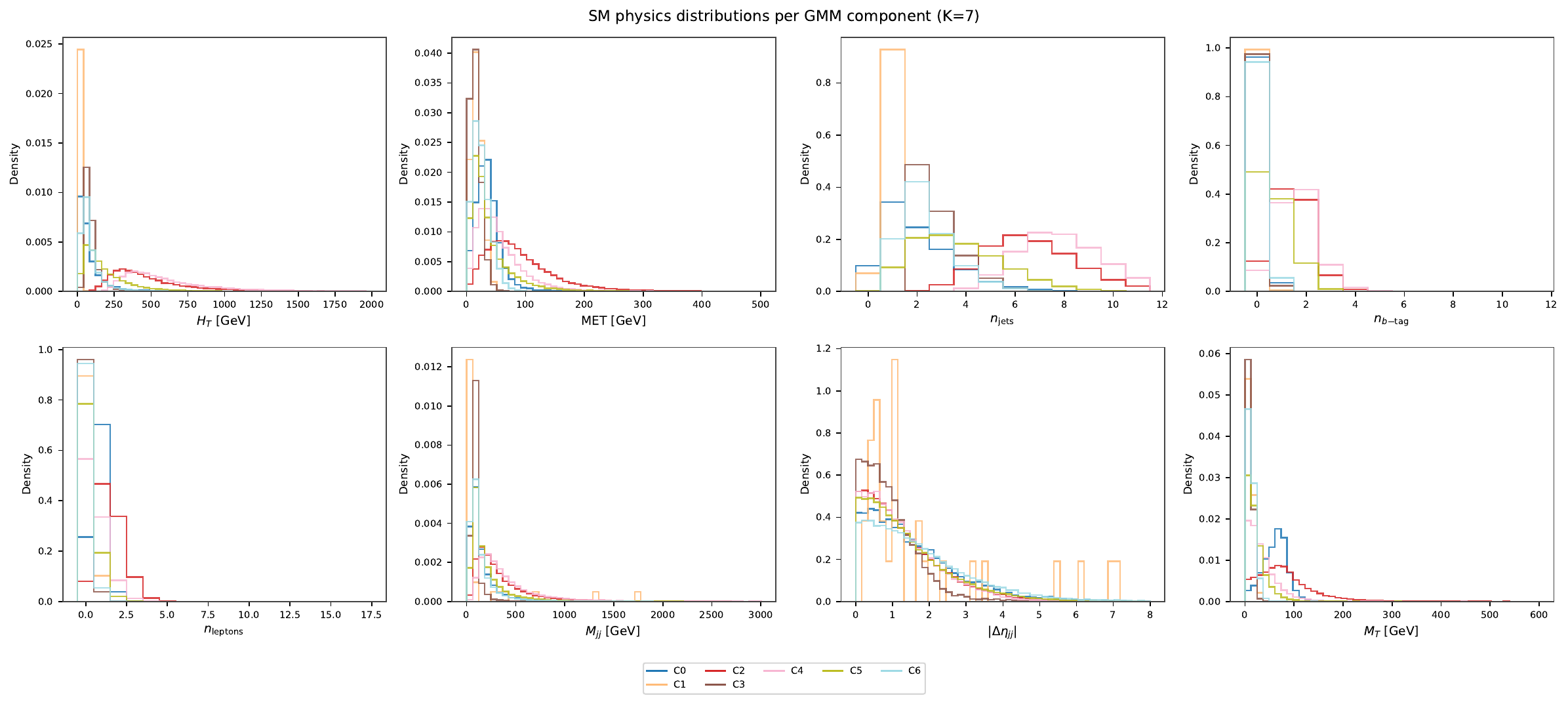}
\caption{Distributions of the eight observables over the Standard Model events of
each mixture component, with the seven components overlaid. This is the material
behind the median-based summary of \cref{fig:gmm_profiles} in the main text, which
reports one number per cell where this reports the full shape. No process labels
enter the construction: the mixture is fitted on the embeddings alone and each
region is described only by the physics of the events assigned to it.}
\label{fig:physics_per_component}
\end{figure}

\cref{fig:physics_per_component} reports the same eight observables over the SM events of each of the $K=7$ mixture components, with the seven components overlaid. Where \cref{fig:gmm_profiles} summarizes each region by a single standardized median deviation, this figure shows the full shapes from which those medians are computed.

\clearpage
\end{appendix}

\bibliography{SciPost_Example_BiBTeX_File.bib}

\end{document}